\PassOptionsToPackage{dvipsnames}{xcolor}
\documentclass[notoc]{kling-keye}
\ifdefined
\fi
\usepackage[shortlabels,inline]{enumitem}
\definecolor{cvprblue}{rgb}{0.21,0.49,0.74}
\usepackage{xspace}
\usepackage{multirow}
\usepackage{pifont}
\usepackage{tabularx}
\usepackage[most]{tcolorbox}
\newcommand{\ours}{VibePrevisBench\xspace}
\newcommand{\cmark}{\textcolor{ForestGreen}{\ding{51}}}
\newcommand{\xmark}{\textcolor{BrickRed}{\ding{55}}}
\newcommand{\pmark}{\textcolor{Orange}{$\circ$}}
\definecolor{rulec}{HTML}{2A78D6}
\definecolor{vlmc}{HTML}{EB6834}

\newtcolorbox{observation}[1]{enhanced,colback=rulec!5,colframe=rulec!70!black,boxrule=0.5pt,arc=2pt,
  left=3pt,right=3pt,top=2pt,bottom=2pt,fontupper=\small,title={\small\textbf{Observation #1}},
  coltitle=black,colbacktitle=rulec!15,attach title to upper={\ }, before skip=4pt, after skip=4pt}

\newtcolorbox{takeaway}{enhanced,colback=ForestGreen!4,colframe=ForestGreen!55!black,boxrule=0.5pt,
  arc=2pt,left=4pt,right=4pt,top=3pt,bottom=3pt,fontupper=\small,
  title={\small\textbf{What to fix next}},coltitle=black,colbacktitle=ForestGreen!14,
  before skip=5pt,after skip=5pt}
\usepackage{colortbl}

\newcommand{\agenticon}[1]{\csname agenticon#1\endcsname}

\usepackage{wrapfig}
\usepackage{placeins}
\usepackage{needspace}

\title{VibePrevisBench: Evaluating Agentic Coding\\for 4D Scene Generation in Blender}
\author{{\small Yuzhe Chen$^{*\S}$\quad Muxi Diao$^{*}$\quad Lele Yang\quad Wenbin Xu\quad Xuyang Liu\quad Qile Su\\
Changyi Liu$^{\dagger}$\quad Jiankang Chen\quad Kaiyu Jiang\quad Haoyu Zhang\quad Hongyi Fu\\
Jianxiong Wang\quad Tianke Zhang\quad Haonan Fan\quad Yingxin Li\quad Bingyi Li\quad Xu Wang\\
Xuanyu Zheng$^{\dagger}$\quad Bin Wen$^{\ddagger}$\quad Tingting Gao\quad Jiale Cao$^{\ddagger}$}}
\affiliation{KlingAI Research\quad Zhongguancun Academy\quad The Hong Kong Polytechnic University\\[0.5em]
{\footnotesize $^{*}$Equal contribution.\quad $^{\dagger}$Project lead.\quad $^{\ddagger}$Corresponding author.\\
$^{\S}$This work was conducted during the author's internship at KlingAI.}}
\headertext{VibePrevisBench: 4D Scene Generation in Blender}
\reportdate{November 11, 2026}
\projectlinks{\href{https://vibeprevisbench.github.io/}{vibeprevisbench.github.io}}
\reportabstract{Controllable real-world video generation models can condition on a rendered graybox proxy video, which fixes layout, motion and camera. 
Coding agents can build such videos by generating a 4D scene in Blender, yet no benchmark checks both these rendered proxy videos and the complex, multi-shot scenes behind them.
We introduce \textbf{\ours}, a benchmark for 4D scene generation: an agent turns a brief into an executable 4D scene and renders its proxy video.
Its 242 tasks span four input kinds, including scripts, photos, continuations, and reference videos.
Each delivery is scored along two evaluation dimensions.
\textbf{Rule-based Scene Check} re-runs the code and compares the exported 4D scene with hidden ground truth facts in nine dimensions.
\textbf{VLM-based Screen Check} asks a VLM judge whether facts that pass the rules are recognizable in the proxy video.
We analyze the evaluation results and trajectories of 11 frontier coding agents and identify three key observations.
First, multimodal understanding separates the agents, especially in understanding space and camera motion.
Second, all agents seem to already master basic camera work and staging, and fail at challenging requests.
Third, agents inspect their own renders but fail to correct their errors.
Agents thus need better visual understanding of space and camera, explicit checks for challenging requests, and item-by-item verification of renders against the brief.
}

\begin{document}
\maketitle
\reportcontents

\begin{figure}[htbp]
  \centering
  \includegraphics[width=\linewidth]{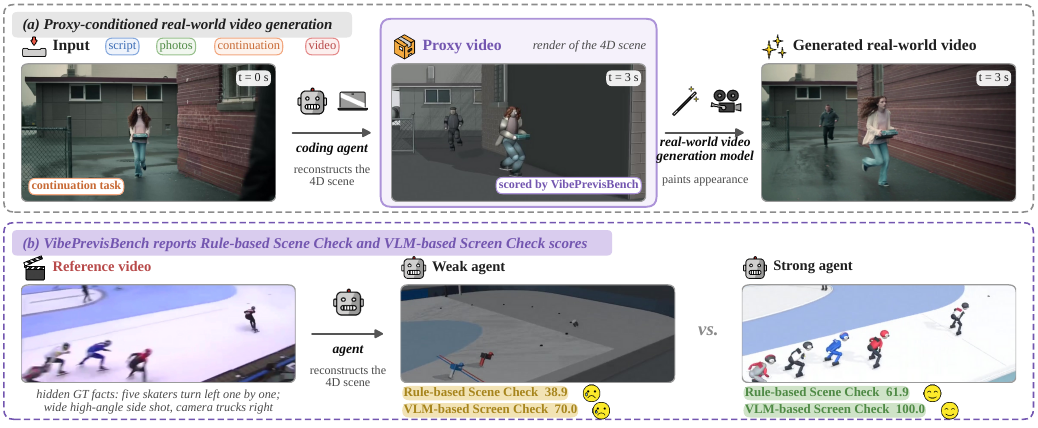}
  \caption{\textbf{Proxy-conditioned real-world video generation and what \ours evaluates.} (a)~From a continuation brief (first frame), a coding agent generates a 4D scene and renders its graybox proxy video on which a controllable real-world video generation model paints appearance~\cite{kling40,seedance25}. (b)~\ours scores Rule-based Scene Check of 4D scene and VLM-based Screen Check of its proxy video against hidden ground truth facts.}
  \label{fig:teaser}
\end{figure}

\begin{figure*}[t]
  \centering
  \includegraphics[width=1.0\linewidth]{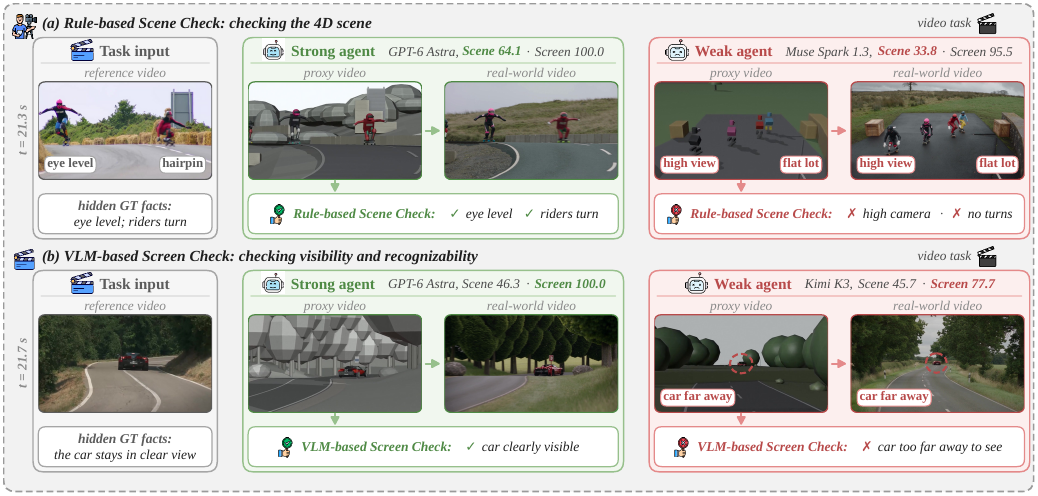}
  \caption{\textbf{Two complementary evaluation dimensions.} Each panel compares a reference video with strong and weak agents' proxy videos and the corresponding real-world videos generated by Seedance~2.5. (a)~Rule-based Scene Check evaluates the 4D scene, and detects the weak agent's high camera and missed turns. (b)~VLM-based Screen Check assesses visibility and recognizability: the weak agent's car passes the in-frame and turning checks but is too distant to recognize clearly. The real-world videos inherit their proxy videos' errors. In addition, we note that the generated real-world videos keep the errors of their proxy videos, demonstrating the importance of proxy video generation quality.}  \label{fig:proxy}
\end{figure*}

\begin{table*}[t]
\centering
\caption{\textbf{Comparison with related benchmarks.} \emph{Staging}: the scene has people, interaction between people or with objects, and actions at given times. \emph{Filming}: camera placement or motion is specified, and several shots are joined by cuts. \emph{Evaluation}: the model generates a \emph{4D scene} that changes over time; it is checked item by item \emph{vs.\ GT}; the frames are scored \emph{on screen}; \emph{checks/task} counts scored items per task. \emph{Duration}: target length in seconds, the median for ours. \cmark~yes, \pmark~partial, \xmark~no, --~not reported. BVB reuses the videos and questions of VSI-Bench. Only \ours is \cmark\ in every column.}
\label{tab:compare}
\small
\setlength{\tabcolsep}{2.0pt}
\resizebox{\linewidth}{!}{\begin{tabular}{@{}l|ccc|r|rccc|cc|cccr@{}}
\toprule
 & \multicolumn{3}{c|}{\textbf{Input}} & & \multicolumn{4}{c|}{\textbf{Staging}} & \multicolumn{2}{c|}{\textbf{Filming}} & \multicolumn{4}{c}{\textbf{Evaluation}} \\
Benchmark & Text & Image & Video & \#Tasks & Duration (s) & People & Interaction & Timed actions & Camera & Multi-shot & 4D scene & vs.\ GT & On screen & Checks/task \\
\midrule
\multicolumn{15}{@{}l}{\scriptsize\textit{Real-world video generation}} \\
VBench-2.0~\cite{zheng2025vbench2} & \cmark & \xmark & \xmark & -- & -- & \cmark & \pmark & \xmark & \pmark & \pmark & \xmark & \xmark & \cmark & -- \\
WorldScore~\cite{duan2025worldscore} & \cmark & \cmark & \xmark & 3k & -- & \xmark & \xmark & \xmark & \cmark & \xmark & \xmark & \pmark & \cmark & -- \\
FilmBench~\cite{wang2026filmbench} & \cmark & \cmark & \xmark & 1.2k & -- & \cmark & \pmark & \xmark & \pmark & \cmark & \xmark & \xmark & \cmark & -- \\
PROWBench~\cite{huang2026prowbench} & \cmark & \pmark & \cmark & 170 & 5 & \cmark & \cmark & \cmark & \cmark & \xmark & \xmark & \pmark & \cmark & -- \\
CodeVideoBench~\cite{liang2026codevideomodel} & \cmark & \cmark & \cmark & 22 & 5 & \pmark & \pmark & \xmark & \cmark & \xmark & \xmark & \xmark & \cmark & -- \\
\midrule
\multicolumn{15}{@{}l}{\scriptsize\textit{Video understanding}} \\
VSI-Bench~\cite{yang2024vsibench} & \xmark & \xmark & \cmark & 288 & $\sim$100 & \xmark & \xmark & \pmark & \xmark & \xmark & \xmark & \xmark & \xmark & 18 \\
CameraBench~\cite{lin2025camerabench} & \xmark & \xmark & \cmark & 3k & 5.7 & \pmark & \xmark & \xmark & \cmark & \xmark & \xmark & \xmark & \xmark & -- \\
OmniCompass~\cite{liu2026omnicompass} & \xmark & \pmark & \cmark & 820 & 354 & \cmark & \pmark & \cmark & \xmark & \xmark & \xmark & \xmark & \xmark & 3 \\
\midrule
\multicolumn{15}{@{}l}{\scriptsize\textit{Video-conditioned scene generation}} \\
4DCodeBench~\cite{shen2026_4dcodebench} & \xmark & \xmark & \cmark & 200 & $\sim$7 & \xmark & \xmark & \pmark & \xmark & \xmark & \cmark & \pmark & \cmark & 6 \\
BVB~\cite{tang2026bvb} & \xmark & \xmark & \cmark & 288 & $\sim$100 & \xmark & \xmark & \xmark & \pmark & \xmark & \pmark & \xmark & \cmark & 18 \\
CineRecon~\cite{wang2026cinerecon} & \xmark & \xmark & \cmark & 50 & 15 & \cmark & \cmark & \cmark & \cmark & \pmark & \cmark & \xmark & \cmark & 193--400 \\
\midrule
\multicolumn{15}{@{}l}{\scriptsize\textit{Scene generation in code}} \\
3DCodeBench~\cite{gao2026_3dcodebench} & \cmark & \cmark & \xmark & 212 & -- & \xmark & \xmark & \xmark & \xmark & \xmark & \pmark & \cmark & \cmark & -- \\
BlenderWorld~\cite{fang2026blenderworld} & \cmark & \cmark & \cmark & 223 & -- & \pmark & \pmark & \xmark & \pmark & \xmark & \pmark & \pmark & \cmark & 12 \\
Code4Scene~\cite{ye2026code4scene} & \cmark & \cmark & \xmark & 190 & -- & \xmark & \xmark & \xmark & \xmark & \xmark & \pmark & \cmark & \cmark & -- \\
CinemaBench~\cite{zhang2026filmflow} & \cmark & \xmark & \cmark & 810 & $\sim$10 & \pmark & \xmark & \xmark & \cmark & \xmark & \xmark & \xmark & \cmark & -- \\
\midrule
\rowcolor{cvprblue!8}\textbf{\ours (ours)} & \cmark & \cmark & \cmark & 242 & 40 & \cmark & \cmark & \cmark & \cmark & \cmark & \cmark & \cmark & \cmark & \textbf{408} \\
\bottomrule
\end{tabular}}
\end{table*}

\section{Introduction}
\label{sec:intro}

{\emergencystretch=1em
Controllable real-world video generation models now use previsualization (previs), a rough 3D mock-up of a shot, to guide real-world video generation~\cite{Wong2012,Muender2018,kling40,seedance25}. This input is a graybox proxy video, the untextured render of a 4D scene whose layout, people, actions, cameras and cuts unfold over time. Conditioning on proxy video makes all of these more controllable~\cite{jiang2025vace,gu2025das,nvidia2025cosmostransfer1,cai2026liver,liang2026codevideomodel}, but building it by hand takes skilled artists and time~\cite{hu2026previzwhiz,yin2026stateflow}. Coding agents can generate the scene instead: from a script, photos, a continuation or real footage, they write Blender code, render previews, inspect their frames and revise~\cite{li2026videococo,chen2026wnm,chen2026codeworldmodel,liang2026codevideomodel,li2026storyblender,diao2026vibevideobench}. We call this task 4D scene generation: the agent writes the executable 4D scene that a brief describes or shows, and renders its proxy video. \Cref{fig:teaser}(a) shows this pipeline. \emph{The agent becomes the director; the real-world video generation model, the renderer.}\par}

The proxy video shapes the final real-world video: the real-world video generation model keeps the camera and layout of proxy video, so an agent's errors carry into the resulting real-world video, as \cref{fig:proxy} shows. Therefore, it is important to generate the high-quality proxy video. To evaluate the quality of proxy video, we need to check two things. First, is the 4D scene right? In \cref{fig:proxy}(a), the task asks for riders turning at eye level, but the weak agent films them from high above a flat lot. The proxy video looks fine, and the VLM judge even gives it 95.5. Thus, it is necessary to find this error by rule-based scene check in 4D scene. Second, can a viewer see it is right? In \cref{fig:proxy}(b), the car is in frame and turning, so the rules pass it. However, the car is a tiny dot on screen, which can be found by the VLM-based screen check.

{
Existing benchmarks support such an evaluation only in part. \Cref{tab:compare} compares them along four dimensions and shows three gaps.
\textbf{(i) Narrow inputs.} Benchmarks of dynamic scenes mostly start from text prompts~\cite{zheng2025vbench2,wang2026filmbench} or reference videos~\cite{shen2026_4dcodebench,tang2026bvb,wang2026cinerecon}. A director, in contrast, may start from a script, photos of a real location or footage to restage, where each source tests a different skill.
\textbf{(ii) Static content.} Benchmarks for scene generation mostly target single objects, static scenes or short clips without people~\cite{gu2025blendergym,gao2026_3dcodebench,fang2026blenderworld,ye2026code4scene,shen2026_4dcodebench}. None tests a long, multi-shot scene in which several people's routes, event timing, camera moves and cuts must hold together.
\textbf{(iii) Screen-only evaluation.} Dynamic-scene benchmarks score only proxy or real-world video frames~\cite{wang2026filmbench,wang2026cinerecon}, while checks of scene state against ground truth cover only static scenes or short clips~\cite{gao2026_3dcodebench,ye2026code4scene,shen2026_4dcodebench}.\par}

To address these gaps, we introduce \textbf{\ours}, a benchmark for 4D scene generation: an agent turns a brief into an executable 4D scene and renders its proxy video. \Cref{fig:teaser}(b) shows what it scores: the 4D scene and its proxy video, both checked against hidden GT facts. \ours has three key features that match the three gaps mentioned above.
\textbf{(i) Diverse inputs.} Briefs come in four kinds, scripts, photos, continuations and reference videos, across seven scene domains, as \cref{fig:stats} shows.
\textbf{(ii) Staged and filmed scenes.} Most tasks stage several people in multi-shot scenes of about 40~s, so routes, event timing, camera moves and cuts are tested together.
\textbf{(iii) Two evaluation dimensions.} \textbf{Rule-based Scene Check} re-runs the delivered code, exports the 4D scene frame by frame and checks it against hidden GT facts in nine dimensions. \textbf{VLM-based Screen Check} asks a VLM judge whether a viewer can recognize each entity, read each action and follow the story in the proxy video.

We evaluate 11 frontier coding agents, and we make three observations. First, multimodal understanding separates the agents, especially in understanding space and camera motion. The gap between the best and the worst agent grows from 12.0 Rule-based Scene Check points on scripts to 19.9 on photos, while weak agents require more rounds facing failed API calls. Second, all agents master only basic camera work and staging, and fail at challenging requests. When the reference films from a low angle, 66\% of shots come out at eye level, whereas eye-level shots go wrong only 9\% of the time. Third, agents inspect their own renders but fail to correct their errors. Even when agents say their renders match the reference, camera-angle errors are just as frequent as in runs without such claims. These observations call for better visual understanding of space and camera, explicit checks for challenging requests, and item-by-item verification of renders against the brief. In summary, our contributions are:

\begin{itemize}[leftmargin=*,itemsep=0pt,topsep=2pt]
 \item \textbf{A benchmark for 4D scene generation.} \ours has 242 tasks over four input kinds: scripts, photos, continuations and reference videos, in seven scene domains. Most tasks stage several people in long, multi-shot scenes to jointly test routes, event timing, camera motion and cuts.
 \item \textbf{Two evaluation dimensions.} Rule-based Scene Check tests whether the 4D scene is right by checking it against hidden GT facts, and VLM-based Screen Check tests whether what is right can be seen in the proxy video. 
 \item {\emergencystretch=2em
 \textbf{A diagnostic study of 11 coding agents.} From their scores, trajectories and ablations, three observations show what separates the agents, where they all fail and why they do not correct themselves.\par}
\end{itemize}

\section{Related Work}
\label{sec:related}
\noindent\textbf{Proxy-conditioned real-world video generation.}
Controllable real-world video generation models use camera paths to control viewpoints~\cite{he2024cameractrl,wang2024motionctrl,bai2025recammaster} and object trajectories to control motion~\cite{fu2025_3dtrajmaster,wang2025cinemaster,zou2026miseenvideo}. A proxy video rendered from a coarse 4D scene can control layout, motion and camera together~\cite{gu2025das,ren2025gen3c,yang2026coag,zhou2026proxyup,jiang2025vace,nvidia2025cosmostransfer1,cai2026liver}. Coding agents can generate the 4D scene in code and render its proxy video~\cite{hu2026previzwhiz,duan2026pace,yin2026stateflow,li2026storyblender,nan2026mindofdirector}. Controllable real-world video generation models use this proxy to generate real-world video and may inherit its scene errors. It is therefore important to evaluate proxy video quality. Yet existing video generation benchmarks mainly evaluate the final video. General benchmarks assess visual quality, temporal consistency and alignment with the input~\cite{huang2024vbench,zheng2025vbench2,sun2025t2vcompbench,duan2025worldscore,meng2024phygenbench,bansal2025videophy2,li2025worldmodelbench,wang2025videoverse}; specialized benchmarks use simulation references or physical constraints to assess object and human motion~\cite{jain2026rigidbench,chen2026pvbench,fang2026humanscore}. Our focus is the upstream coding agent: whether its 4D scene meets the brief and whether the required content can be clearly recognized in its proxy video.

\noindent\textbf{Understanding a video through scene generation.}
Video understanding benchmarks assess perception through answers, groundings or captions~\cite{fu2024videomme,yang2024vsibench,meng2026videozerobench,liu2026omnicompass,chai2024auroracap,wang2024tarsier,yao2026timechatcaptioner}; camera benchmarks reveal persistent errors in camera-motion understanding~\cite{lin2025camerabench,cheng2026leftisright}.
Correct answers do not guarantee a coherent scene representation~\cite{liu2026spatialbabel,huang2026camo}.
Generating a video's 4D scene in code makes objects, motion and cameras explicit. Existing benchmarks cover physics clips, indoor scans, embodied demonstrations and short cinematic clips~\cite{shen2026_4dcodebench,tang2026bvb,tang2026video2world,wang2026cinerecon}.
None combines multi-shot scene generation with GT checks of the generated scene (\cref{tab:compare}).

\noindent\textbf{Generating scenes with code.}
LLM agents generate objects, scenes, animations and films through Blender code~\cite{sun2023threedgpt,hu2024scenecraft,huang2024blenderalchemy,yin2026viga,he2024kubrick,xu2025filmagent}. Code world models also generate 4D scenes whose proxy videos condition real-world video generation~\cite{li2026videococo,chen2026wnm,chen2026codeworldmodel,liang2026codevideomodel,fang2026coder}.
Agent benchmarks mostly cover edits, objects or static scenes~\cite{gu2025blendergym,gao2026_3dcodebench,fang2026blenderworld,ye2026code4scene,zhao2026sceneactbench}; camera benchmarks assume a given scene~\cite{zhang2026filmflow,li2026cinematraj}.
These benchmarks do not jointly test spatial layout, event timing, object and character motion, and camera motion across multiple shots.

\Cref{tab:compare} summarizes these gaps in inputs, scene complexity and evaluation. \ours evaluates agents that generate executable 4D scenes from scripts, photos, continuations or reference videos and render their proxy videos. Rule-based Scene Check compares the scene with hidden GT facts. VLM-based Screen Check asks whether the facts that pass are recognizable in the proxy video.

\section{\texorpdfstring{\ours}{VibePrevisBench}}
\label{sec:bench}

\begin{figure*}[t]
  \centering
  \includegraphics[width=1.0\linewidth]{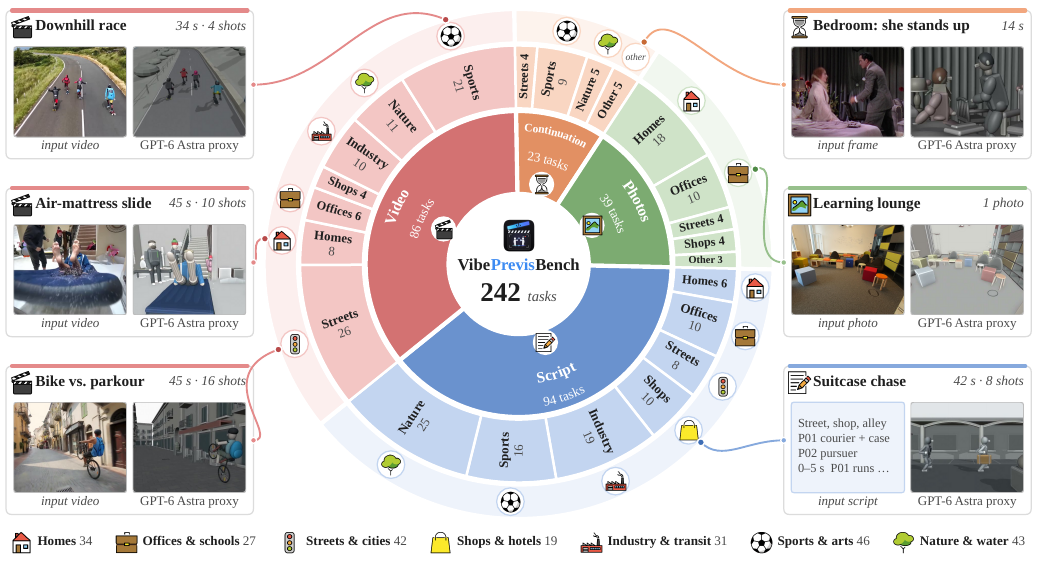}
  \captionsetup{justification=raggedright,singlelinecheck=false}
  \caption{\textbf{Overview of \ours.} The inner ring groups the 242 tasks by input kind: scripts, photos, continuations and reference videos. The outer ring shows their distribution across seven scene domains; arc lengths are proportional to task counts. Each example card pairs a task input with a frame from GPT-6 Astra's generated proxy video.}
  \label{fig:stats}
\end{figure*}

\ours contains 242 tasks that ask agents to generate executable 4D scenes and render proxy videos from briefs. Each brief is paired with hidden GT facts specifying the scene requirements. The tasks cover four input kinds and seven scene domains (\cref{fig:stats}).

\subsection{Task formulation}
\label{sec:task}
\noindent\textbf{Briefs and deliveries.} A task $\tau=(b,m,E,o)$ gives an English brief $b$, optional media $m$, a list $E$ of places and subjects with ids, and output requirements $o$: the frame count $T$, frame rate and resolution of the proxy video. The input kinds differ only in what specifies the scene. A script describes it in prose alone, without distances or angles. Photo and reference video tasks show it in $m$, as one to five photos or a clip, while $b$ only names the entries of $E$. A continuation shows the opening frame of a clip and describes the next seconds. The agent answers with a Blender program $p$. To score it, we run $p$ and read out the 4D scene $S=(\{x_e(t)\}_{e\in E},\,c(t),\,\mathcal{K})$: for every place and subject listed in $E$, its oriented box $x_e(t)$ at each frame $t=1,\dots,T$, together with the camera $c(t)$ and the cut times $\mathcal{K}$. Rendering $S$ gives the proxy video $V$, with the same $T$ frames.

\noindent\textbf{Rubric.} The hidden GT facts form a rubric $F=\{f_j\}$, where each $f_j$ is one checkable statement about the scene. The facts state which subjects exist, where they stand and face, how the places connect, who moves where and when, and how each camera frames, moves and cuts. For example, a fact may require that P01 stops between 3.9 and 4.9~s, within a 1.5~s tolerance. Its score $s_j(S)\in[0,1]$ measures how well the agent's scene $S$ meets $f_j$. A fact met within its tolerance scores 1, and a metric fact loses credit linearly once its error exceeds the tolerance. All tolerances are fixed before any agent is run. Each fact belongs to one of nine dimensions: entities, layout, spaces, picture structure, routes, events, camera, cuts and physical validity. Picture structure compares the depth, normals and segmentation of each shot with the reference frames, so only photo, continuation and video tasks have it; Appendix \cref{tab:dims} defines all nine. Appearance is never checked, since the real-world video generation model supplies it. Whatever the input kind, the agent delivers the same kind of program $p$, which is scored through the same $S$ and $V$ against facts of the same form.

\subsection{Rule-based Scene Check and VLM-based Screen Check}
\label{sec:eval}

\begin{figure*}[t]
  \centering
  \includegraphics[width=1.0\linewidth]{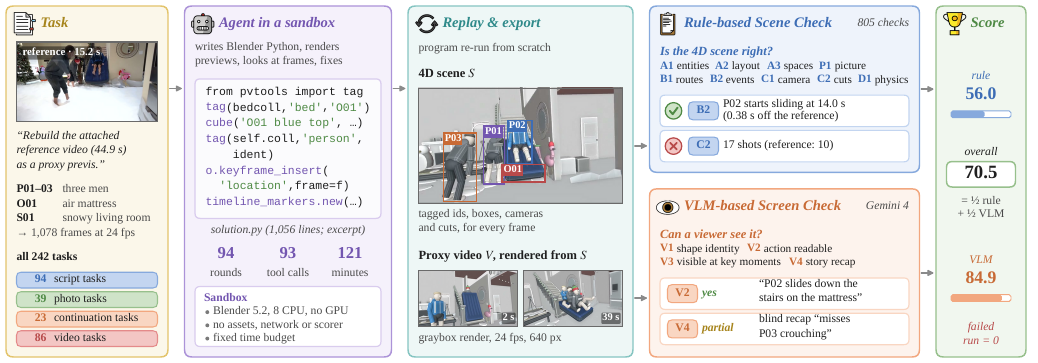}
  \caption{\textbf{One task through the evaluation pipeline.} The agent writes Blender Python in a sandbox and inspects its own renders. We re-run the delivered program from scratch. The replay exports the 4D scene $S$, the tagged boxes, cameras and cuts of every frame, and renders the proxy video $V$ from it. Rule-based Scene Check compares $S$ with the hidden GT facts, two of 805 checks shown; VLM-based Screen Check asks the VLM judge about $V$, only for items the rules passed, two verdicts shown.}
  \label{fig:pipeline}\label{fig:validity}
\end{figure*}

Rule-based Scene Check asks whether the scene is right, and VLM-based Screen Check whether what is right can be seen. \Cref{fig:pipeline} follows one task through both, and \cref{sec:validity} validates them with injected defects, re-scoring, three VLM judges and a human study.

\noindent\textbf{Delivery and replay.} Each delivery must pass three gates: its program runs on its own, renders the full video and reopens in Blender. A delivery that fails a gate scores 0 on both checks. Such failures stay in the average, so an agent gains nothing by failing hard tasks. Every other program is re-run from scratch in a fresh sandbox, which exports $S$ frame by frame and renders $V$. Both checks use only the replay outputs, so the evaluated scene and video come from the same program execution.

\noindent\textbf{Rule-based Scene Check.} It binds subjects and places by their tagged ids and matches untagged objects by class. Script facts are relations, since scripts give no coordinates, whereas media facts are metric and are compared after each place is aligned to the ground truth by a similarity transform. The score is then built in three steps. First, the facts of one kind form a block $b$, whose raw score $P_b$ is the mean of $s_j$ over its facts. Second, each block is corrected for chance and for missing entities: $Q_b=\mathrm{clip}\big((P_b-R_b)/(1-R_b)\big)\,\kappa_b$, where clip bounds the value to $[0,1]$. The chance rate $R_b$ is the $P_b$ that the agent's own objects earn when shuffled at random within each place, and $\kappa_b$ is the share of the block's facts whose entities exist. A random layout thus scores 0, extra clutter raises $R_b$ along with $P_b$, and every missing entity lowers $\kappa_b$. Third, blocks combine into dimension scores $Q_d$ with weights fixed in advance, and $\mathrm{Scene}=100\cdot\frac{1}{|D_\tau|}\sum_{d\in D_\tau}Q_d$ averages them over the dimensions $D_\tau$ that the task exercises. The scorer is deterministic, and \cref{sec:appscoring} details the scoring.

\noindent\textbf{VLM-based Screen Check.} A VLM judge, Gemini~4, watches the proxy video at 2~fps, plus an orbit render of the delivered scene on photo tasks, blind to the agent's name and the object names. Without the brief, it first retells the clip. It then receives the brief, the reference video on video tasks, and checklists from the hidden GT facts, limited to the items the rules passed, and rates each item as no, partial or yes, $v_{k,i}\in\{0,\frac12,1\}$. The items fall into four dimensions: recognition, action, visibility and story. Recognition counts as yes only if the blind retelling already named the subject or place, and story asks whether the retelling recounts the story beats. With $\bar v_k$ the mean verdict of dimension $k$, $\mathrm{Screen}=100\cdot\frac{1}{|K_\tau|}\sum_{k\in K_\tau}\bar v_k$, where photo tasks use recognition only, and the overall score is $(\mathrm{Scene}+\mathrm{Screen})/2$. \Cref{sec:appjudge} gives the judge's rubric.

\begin{figure*}[t]
  \centering
  \includegraphics[width=0.95\linewidth]{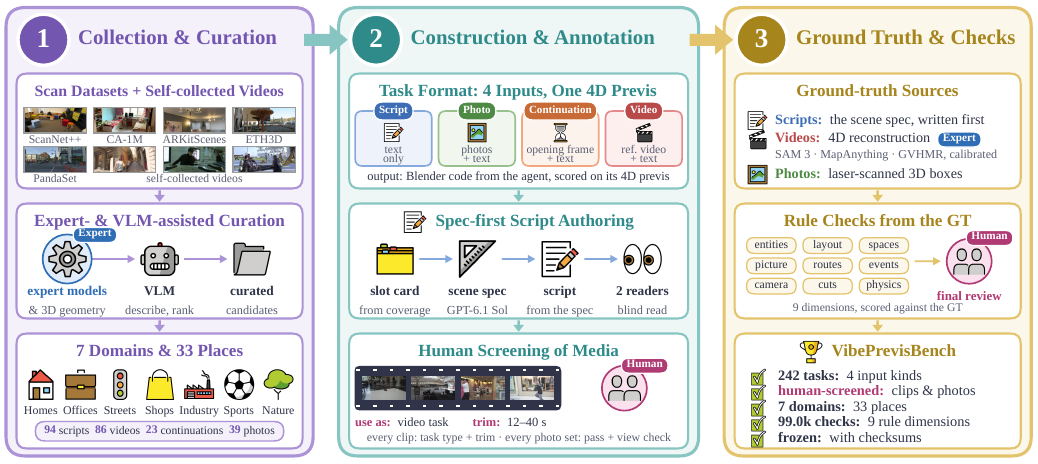}
  \captionsetup{justification=raggedright,singlelinecheck=false}
  \caption{\textbf{How \ours is built.} (1)~Scan datasets and candidate videos are curated with expert models and a VLM into seven scene domains and 33 places. (2)~Scripts are written spec-first and must survive two blind readers; humans select and trim clips and screen photo sets. (3)~Hidden GT facts come from the scene spec, 4D reconstruction or scanned 3D boxes, and humans review the resulting checks in nine dimensions.}
  \label{fig:construct}
\end{figure*}

\subsection{Construction}
\label{sec:construct}
We first curate source material, then create task briefs, and finally derive and review the hidden GT facts (\cref{fig:construct}). \Cref{sec:appconstruct} details the coverage taxonomies, selection criteria and task pruning.

\noindent\textbf{Stage 1: Collection and curation.} Two frozen taxonomies, of what a scene holds and of how a task is asked, set quotas that new tasks fill. Clips for video and continuation tasks come from over 27k source videos. Keeping a few clips per source with several moving people, camera motion or cuts leaves 1.8k clips, of which 388 pass VLM checks for live action, visible events and distinct subjects. Human reviewers choose among these, and \cref{sec:appconstruct} describes the steps from these clips to the final video and continuation tasks. Of the 39 photo tasks, 38 draw on scan datasets with measured 3D boxes and camera poses.

\noindent\textbf{Stage 2: Task construction.} Every brief is derived from a fixed scene. For a script, a model builds the scene from a card and then writes the script, each scored claim citing its sentence. Two blind readers from the Claude and DeepSeek families see only the script and write down where everything is and when it happens; a claim that neither recovers is reworded or dropped. A media brief gives each place and subject an id and a description. Human reviewers check and correct every brief and its hidden GT facts.

\noindent\textbf{Stage 3: Ground truth and checks.} Scripts take their hidden GT facts from the built scene, in which every claim is checked to hold, and photo tasks from their scans. For videos, most continuations and one photo task, a pipeline estimates it: it recovers shots, subjects, cameras, depth and bodies and lifts them to 3D, as Appendix \cref{fig:gtvideo} shows. Facts the pipeline cannot verify are marked unscorable and never count. All 11 agents then ran the candidates, and tasks that did not separate them were dropped, judged by item-rest correlation and spread rather than by any single agent's score. Tasks, media, hidden GT facts and scorer are frozen with checksums, so every score can be reproduced from the release.

\begin{figure*}[t]
  \centering
  \includegraphics[width=1.0\linewidth,trim=0 0 0 267.8bp,clip]{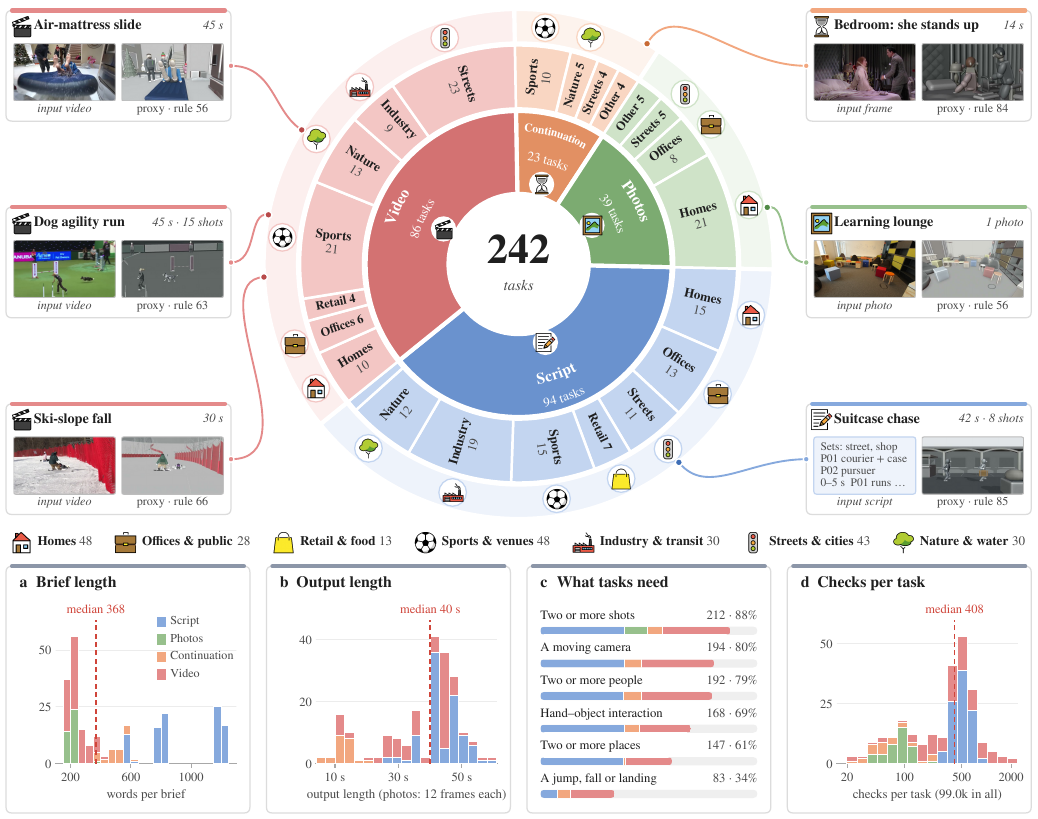}
  \captionsetup{justification=raggedright,singlelinecheck=false}
  \caption{\textbf{Task statistics by input kind.} Dashed lines mark medians. \textbf{(a)}~Brief length; scripts cluster near 600, 800 and 1,200 words. \textbf{(b)}~Output duration; photo tasks display each view for 12 frames, followed by an 8~s camera sweep. \textbf{(c)}~Task counts and proportions for each staging demand. \textbf{(d)}~Rule checks per task on a logarithmic scale; one fact may yield several checks, such as pairwise event orders.}
  \label{fig:statsdist}
\end{figure*}

\subsection{Statistics}
\label{sec:stats}
\noindent\textbf{Coverage and scale.} \ours contains 242 tasks: 94 scripts, 86 reference videos, 39 photo tasks and 23 continuations. As \cref{fig:stats} shows, each of the seven domains holds 19 to 46 tasks and each of the 33 places 3 to 13. In \cref{fig:statsdist}(a), briefs run 153 to 1,200 words with a median of 368: scripts spell out the scene in about 800 words, while media tasks mainly list ids. Outputs run 6 to 60~s with a median of 40~s, as \cref{fig:statsdist}(b) shows.

\noindent\textbf{Staging demands and checks.} \Cref{fig:statsdist}(c) shows that 88\% of tasks have two or more shots, 80\% a moving camera, 79\% two or more people, 69\% hand--object interaction and 61\% two or more places, and 71\% combine several shots, a moving camera and two or more people. A task carries 20 to 2,175 checks, 408 at the median and 99.0k in all, as \cref{fig:statsdist}(d) shows, and exercises a median of eight of the nine dimensions. Events hold 57\% of the checks, since event order is checked pair by pair, but Rule-based Scene Check averages dimensions rather than checks, so they do not dominate the score. \Cref{sec:appstats} splits them by input kind.

\section{Experiments}
\label{sec:exp}

\begin{table*}[t]
\centering
\caption{\textbf{Main results on 242 tasks.} Rule-based Scene Check and VLM-based Screen Check per input kind and over all tasks (\emph{All}); VLM-based Screen Check uses the Gemini~4 judge, which rates only recognition on photos; \emph{Overall} is their mean. Failed runs score 0. Per column, the best score is in \textbf{bold} and the second best is \underline{underlined}. The reasoning effort of each agent is given in gray.}
\label{tab:main}
\newcommand{\agent}[3]{\makebox[0.36cm][c]{\agenticon{#1}}\hspace{0.2em}#2\,{\scriptsize\textcolor{black!45}{#3}}}
\footnotesize
\setlength{\tabcolsep}{3.2pt}
\renewcommand{\arraystretch}{1.12}
\resizebox{\linewidth}{!}{\begin{tabular}{@{}l !{\color{black!20}\vrule width 0.35pt} cccc c !{\color{black!20}\vrule width 0.35pt} cccc c !{\color{black!20}\vrule width 0.35pt} c@{}}
\toprule
\multirow{2}{*}{\textbf{Agent}} & \multicolumn{5}{c!{\color{black!20}\vrule width 0.35pt}}{\textcolor{rulec}{\textbf{Rule-based Scene Check}}} & \multicolumn{5}{c!{\color{black!20}\vrule width 0.35pt}}{\textcolor{vlmc}{\textbf{VLM-based Screen Check}}} & \multirow{2}{*}{\textcolor{ForestGreen}{\textbf{Overall}}} \\
\cmidrule(lr){2-6}\cmidrule(lr){7-11}
 & Script & Photo & Cont. & Video & All & Script & Photo & Cont. & Video & All & \\
\midrule
\agent{openai}{GPT-6 Astra}{max} & \textbf{85.6} & 55.8 & \textbf{79.8} & \textbf{56.8} & \textbf{70.0} & \textbf{76.5} & \underline{93.6} & \textbf{91.1} & \textbf{89.8} & \textbf{85.4} & \textbf{77.7}\\
\agent{openai}{GPT-6.1 Sol}{max} & \underline{84.4} & 55.3 & 78.1 & \underline{56.0} & \underline{69.0} & \underline{73.3} & \textbf{93.8} & \underline{89.5} & \underline{87.3} & \underline{83.1} & \underline{76.1}\\
\agent{gemini}{Gemini 4 Argon}{high} & 82.1 & \underline{57.1} & \underline{78.5} & 55.8 & 68.4 & 72.7 & 90.3 & 88.8 & 85.6 & 81.7 & 75.0\\
\agent{claude}{Opus 5.5}{high} & 83.0 & \textbf{59.8} & 78.2 & 53.9 & 68.4 & 73.0 & 90.4 & 82.5 & 81.6 & 79.8 & 74.1\\
\agent{openai}{GPT-5.6 Sol}{max} & 80.8 & 44.2 & 73.0 & 49.0 & 62.9 & 69.4 & 92.5 & 81.1 & 78.7 & 77.5 & 70.2\\
\agent{deepseek}{DeepSeek V4.1 Flash}{max} & 80.1 & 43.2 & 73.0 & 45.7 & 61.3 & 70.8 & 90.2 & 82.4 & 78.8 & 77.9 & 69.6\\
\agent{kimi}{Kimi K3}{xhigh} & 81.3 & 42.4 & 73.9 & 45.2 & 61.5 & 69.3 & 81.2 & 80.9 & 72.5 & 73.5 & 67.5\\
\agent{gemini}{Gemini 3.8 Flash}{high} & 78.6 & 43.1 & 71.4 & 46.2 & 60.7 & 59.1 & 87.7 & 77.6 & 68.3 & 68.7 & 64.7\\
\agent{zhipu}{GLM 5.3 Flash}{high} & 76.0 & 39.9 & 71.0 & 42.6 & 57.8 & 61.0 & 84.8 & 76.3 & 70.2 & 69.5 & 63.7\\
\agent{bytedance}{Doubao Seed 2.1 Pro}{max} & 73.6 & 40.4 & 65.7 & 42.2 & 56.3 & 56.1 & 86.8 & 69.1 & 67.2 & 66.2 & 61.3\\
\agent{meta}{Muse Spark 1.3}{xhigh} & 77.4 & 40.5 & 67.2 & 40.1 & 57.2 & 58.3 & 84.7 & 72.5 & 61.3 & 65.0 & 61.1\\
\bottomrule
\end{tabular}}
\end{table*}

\subsection{Setup}

\noindent\textbf{Evaluated agents.} We evaluate 11 frontier models as coding agents, each at the reasoning effort given in gray in \cref{tab:main}. The GPT models run in the Codex CLI and all other models in Claude Code. An agent writes and runs Blender code, renders previews, inspects its own frames and revises. A small toolkit tags entities and places with the task's ids, so facts bind to objects without visual matching.

\noindent\textbf{Protocol.} Each run takes place in a sandbox with Blender 5.2, eight CPU cores and no GPU. The agent has a budget of 4~hours and no external assets, network or access to the scorer, so it can rely only on what the brief gives it. Every task is run once, and scores are averaged over tasks with equal weight. Input kinds exercise different dimensions and obtain their hidden GT facts differently, so agents are compared within a kind.

\Needspace{22\baselineskip}
\subsection{Main results}
\suppressfloats[t]
\begin{wrapfigure}{r}{0.62\textwidth}
  \centering
  \includegraphics[width=\linewidth]{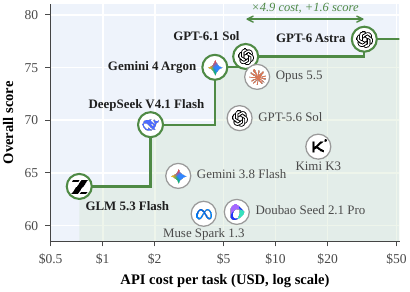}
  \captionsetup{justification=justified,singlelinecheck=false}
  \caption{\textbf{Cost vs.\ score.} Mean task cost at API list prices versus overall score; the staircase marks the cost--score frontier.}
  \label{fig:cost}
\end{wrapfigure}
\Cref{tab:main} reports both checks per input kind; \cref{sec:appprofiles} adds per-dimension profiles. GPT-6 Astra leads with an overall score of 77.7, ahead of GPT-6.1 Sol at 76.1, Gemini 4 Argon at 75.0 and Opus 5.5 at 74.1. Five of the ten adjacent pairs cannot be separated by a paired bootstrap over tasks, while the overall ranking holds across 18 scoring variants in \cref{sec:validity}, at Kendall $\tau \ge 0.85$ with GPT-6 Astra always first. The top four lead the rest by at least 3.9 overall points. GPT-6 Astra leads Rule-based Scene Check on every input kind except photos, where Opus 5.5 leads, as \cref{sec:appwhere} details. On only 35 of the 242 tasks does any agent reach a Rule-based Scene Check score of 90. The remaining gap is shared, not divided: the best of all 11 agents on every task raises the overall score only to 80.6, and the top four alone already reach 80.5.

Cost and score come apart. Per-task cost varies 45-fold, from \$0.73 for GLM 5.3 Flash to \$32.59 for GPT-6 Astra. Five agents sit on the cost--score frontier in \cref{fig:cost}. GPT-6.1 Sol reaches 98\% of the leader's score at 21\% of the cost, and DeepSeek V4.1 Flash 90\% at 6\%. Paying more does not buy rank: Kimi K3 is second in cost but seventh in score.

\begin{figure*}[!b]
  \centering
  \includegraphics[width=1.0\linewidth]{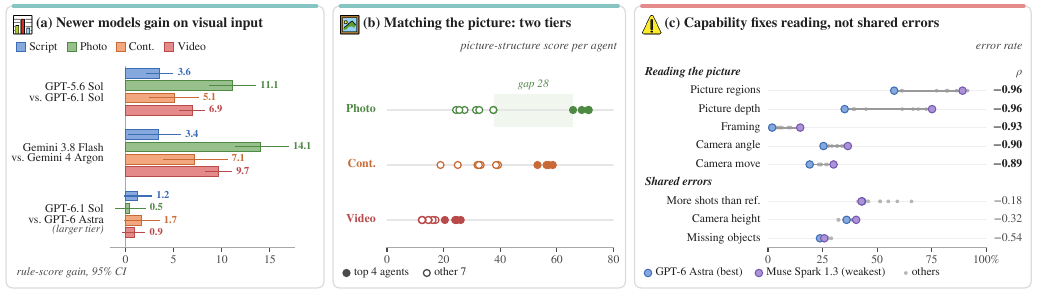}
  \caption{\textbf{Where agents improve, and where they do not.} (a)~Rule-based Scene Check gain of the second model over the first in each pair, per input kind, paired over tasks; whiskers show 95\% bootstrap intervals. The top two pairs are successive generations of one family, while the bottom pair is a step up in tier within one generation at about five times the cost. (b)~Picture-structure score per agent and input kind, which matches each shot's depth, normals and segmentation to the reference frames; filled dots mark the four agents with the highest overall score. (c)~Error rates of the best and the weakest agent, the other nine in gray; $\rho$ is the Spearman correlation with the overall score, and \cref{sec:appwhere} defines each rate.}
  \label{fig:where}
\end{figure*}

\begin{figure*}[t]
  \centering
  \includegraphics[width=1.0\linewidth]{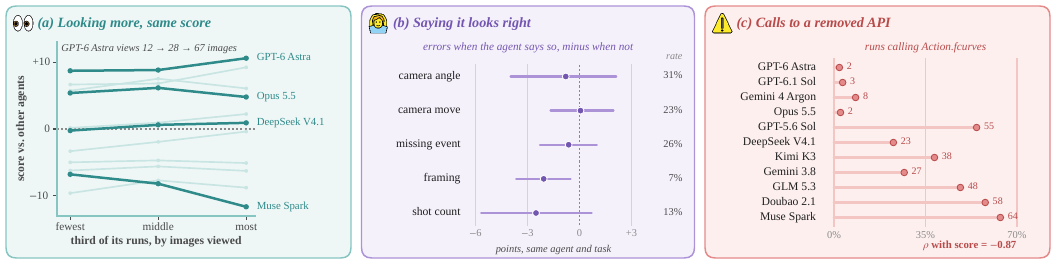}
  \caption{\textbf{How agents look, judge and code.} \textbf{(a)}~Score relative to the other agents' task mean, by third of an agent's runs ranked by images viewed. \textbf{(b)}~Error rate in runs where the agent says its render matches or looks right, minus runs where it does not, with agent and task held fixed; whiskers show 95\% bootstrap intervals and \emph{rate} is the mean error rate. \textbf{(c)}~Share of runs calling \texttt{Action.fcurves}, removed in Blender~5. Rows sorted by overall score.}
  \label{fig:look}\label{fig:habits}
  \vspace{0pt}
\end{figure*}

\subsection{Observations and implications}
We analyze three observations and their implications.

\noindent\textbf{Observation 1: Multimodal understanding separates the agents, especially in understanding space and camera motion.} The more an input relies on vision, the wider the gap between agents. In \cref{tab:main}, the gap between the best and worst agent grows from 12.0 Rule-based Scene Check points on text-only scripts to 19.9 on photos. This gap comes mainly from errors in space and camera. On photos, picture structure and camera account for 63\% of the gap between the top four and the other seven, entities and layout for only 9\%. The top four score 66--71 on picture structure, the other seven 24--38, as \cref{fig:where}(b) shows. The bottleneck is reading images and videos, not building the scene. Given a text description of the hidden GT facts on five video tasks, DeepSeek V4.1 Flash gains 14.0 points, as Appendix \cref{tab:ablation} reports, nearly the full 16.7-point gap between the best and the worst agent on videos. Model updates gain here too: in \cref{fig:where}(a), a newer model of the same family gains 11--14 points on photos but only 3--4 on scripts. A larger tier of the same generation, GPT-6 Astra over GPT-6.1 Sol, gains at most 1.7 points on any input kind. The three weakest agents often call an API removed in Blender~5, as \cref{fig:habits}(c) shows, yet later rounds repair these calls. With agent and task fixed effects, API failures correspond to $+$0.1 points (95\% CI: $-$0.8 to $+$1.0). \emph{Progress on this step thus depends on visual understanding of space and camera, whereas more knowledge of the Blender API saves rounds, not points.}

\noindent\textbf{Observation 2: All agents master only basic camera work and staging, and fail at challenging requests.} Appendix \cref{fig:failures}(a) shows that the errors of all 11 agents concentrate on challenging requests, which they turn into the common choice. Based on camera pitch, 66\% of reference low-angle shots come out at eye level rather than high. Only 4\% of orbits around a subject stay orbits, and 47\% become sideways moves. Half of the video and continuation runs have more shots than the reference, while only 4\% have fewer. Objects mentioned only in prose go missing 35\% of the time. The common version of each request, by contrast, mostly succeeds: eye-level shots go wrong 9\% of the time, follow and push-in moves 5--8\%, and listed objects go missing 11\% of the time. Challenging requests are likely scarce in the data agents learn from, so agents fall back on what they know. The errors lean one way, toward the common choice, whereas a misreading would err both ways. They also barely vary across agents: in \cref{fig:where}(c), errors in reading the picture fall with the overall score, whereas every agent films 32--41\% of shots at the wrong camera height. \emph{Agents therefore need explicit checks for challenging requests, such as camera height and move, shot count and objects named only in prose.}

\noindent\textbf{Observation 3: Agents inspect their own renders but fail to correct their errors.} Agents inspect their renders and view 13--40 images per run, including reference frames, yet more looking does not help. In \cref{fig:look}(a), the third of runs that looks most views 3.3 times as many images as the third that looks least, yet scores only 0.8 points higher at the median once task difficulty is removed. Removing every image from DeepSeek V4.1 Flash's context on eight script tasks changes its score by only $-$0.8 points, as Appendix \cref{tab:ablation} reports. Agents' self-assessments are unreliable. In 1,377 of 2,582 runs with comments after a render, the agent states that its render matches the reference or looks right. With agent and task held fixed, these runs get camera angles, camera moves and events wrong as often as the other runs, as \cref{fig:look}(b) and \cref{sec:appwhere} show. Only framing and shot count, which show at a glance, are 2--3 points lower. \emph{More renders and looking thus do not correct errors; agents need item-by-item verification of renders against the brief.}

\section{Conclusion}
\label{sec:discussion}
We introduce \ours, a benchmark that evaluates both the 4D scenes generated by coding agents and their rendered proxy videos. Its 242 tasks span scripts, photos, continuations and reference videos, covering spatial layout, subject motion, camera motion and cuts. Rule-based Scene Check compares each scene with hidden ground truth across nine dimensions; VLM-based Screen Check tests whether facts that pass are recognizable on screen. Analysis of 11 agents reveals three observations: visual understanding of space and camera separates agents; challenging staging and filming requests remain difficult; and render inspection does not reliably lead to correction. These observations point to three priorities: better visual understanding, explicit checks for challenging requests, and item-by-item verification of renders against the brief. Together, the two checks make scene correctness and on-screen readability measurable.

\clearpage
\bibliographystyle{plainnat}
\bibliography{references}

\clearpage
\section*{Supplementary Material}
\phantomsection
\addcontentsline{toc}{section}{Supplementary Material}
\appendix

\section{Benchmark details}

\subsection{Task construction}
\label{sec:appconstruct}

\paragraph{Coverage first.} Two frozen taxonomies tag every task. One covers what a scene holds: places, spaces, layout relations, subjects, actions, camera and editing, and genre, in 67 categories. The other covers how a task is asked: input kind, writing style, information level, organization and length. Each tag and each kind of place has a minimum number of tasks, and the gaps became slots for new tasks, routed to three lanes.

\paragraph{Scripts.} Each slot becomes a card fixing genre, writing style, length, places, people, actions and a story seed. Scripts come in seven writing styles, from screenplays and shot lists to ad briefs and novel-like prose, at about 600 to 1,200 words. The hidden GT facts are written before the script: a model (GPT-6.1 Sol) first codes the scene, then writes the script, then the claims, each citing the sentence it comes from and checked to hold in the coded scene. Automatic gates check every task's scoring and output specification. Two blind readers from the Claude and DeepSeek families, who see only the script, try to recover every claim; a claim both misread is reworded and re-read up to twice, then dropped if still misread.

\begin{figure*}[!htbp]
  \centering
  \includegraphics[width=\linewidth]{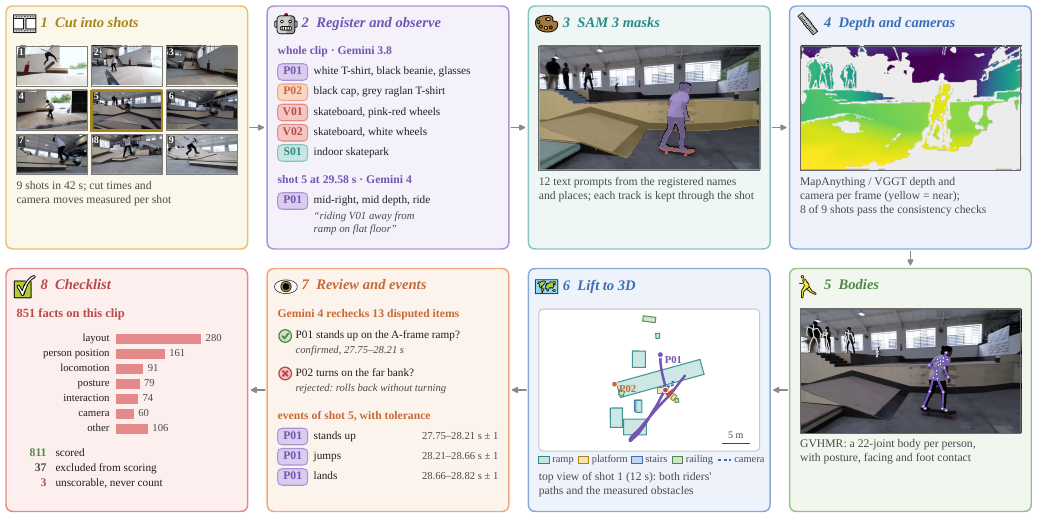}
  \caption{\textbf{Hidden ground truth facts for a video task.} The eight pipeline steps on one clip: shots, registration and per-shot observation, masks, depth and cameras, bodies, lifting to 3D, review and events, and the final checklist of facts with their tolerance windows.}
  \label{fig:gtvideo}
\end{figure*}

\paragraph{Videos and continuations.} We curated candidate video clips. The screening keeps clips of at least 5~s and 2~fps, with a short edge of at least 480 pixels for local clips and 720 for web videos, and drops clips whose adjacent frames are more than 85\% identical. Local clips are cut into windows of 25 to 50~s, a seeded eighth also into 15~s windows, and web videos into 25--45~s video windows and 12~s continuation windows; each source keeps at most three or four windows that overlap by at most 60\%. Windows are ranked by subject presence, the number of moving people or vehicles and its changes, motion beyond the camera's own, camera motion, cuts, person size and duration, and flagged for letterboxing, burned-in text, repeated frames and darkness. We then pre-screened the windows by model ratings of contact sheets (Gemini~4, Claude Sonnet~5.5), and tagged the survivors by place, genre, people, shots and moving vehicles. Human reviewers decided whether a clip becomes a video task, a continuation or neither. A video brief gives an id and a description for each place and subject, and the output must start where the clip starts and cut where it cuts. A continuation brief gives the clip's opening frame and an account of the next 6 to 20~s, mostly following the clip. \Cref{fig:gtvideo} shows how one clip's hidden GT facts are built.

\paragraph{Photos.} We drew on scan datasets with measured 3D boxes and camera poses, indoors and in street scenes, and used each scene once. Boxes from the hidden GT facts were drawn back onto every photo and checked first by a model and then by a human; candidates whose boxes did not match the photo were rejected. The output must show each photo's view for 12 frames and then sweep the place for 8~s. One photo task instead uses five keyframes of a real clip, and its hidden GT facts are estimated as for a video.

\paragraph{Pruning.} All 11 agents ran the pool, and we dropped tasks that did not discriminate between agents: low item-rest correlation or little spread across agents, never a single agent's score. Finally we removed tasks with source or licensing problems or an unscorable run.

\paragraph{Model assistance and fairness.} Several of the evaluated models help build tasks. GPT-6.1 Sol writes the script specs and scripts; the blind readers come from other families, Claude and DeepSeek; Claude Sonnet~5.5, of the same family as Opus~5.5, pre-screens clips; and Gemini~3.8 and Gemini~4 register and observe the subjects behind the hidden GT facts of videos and continuations, while Gemini~4 also rates web clips. Three safeguards limit their influence. Agents are scored by rule checks on facts, never by similarity to a model's own output; the hidden GT facts are never edited to make a reader pass; and every brief and its facts are reviewed and corrected by hand. Ranked by Rule-based Scene Check, the agents show no strong self-preference. GPT-6.1 Sol is second on scripts and over all tasks; DeepSeek V4.1 Flash is seventh on scripts and over all tasks; and Opus~5.5 is third over all tasks and fourth on videos. The Gemini models rank somewhat higher on the media tasks whose hidden GT facts they helped estimate: Gemini~4 Argon is fourth over all tasks but third on videos and second on continuations, and Gemini~3.8 Flash is eighth over all tasks but sixth on videos. A small preference cannot be ruled out; the judge's own-family bias is measured separately in \cref{sec:validity}.

\subsection{Task examples}
\begin{figure*}[!htbp]
  \centering
  \includegraphics[width=\linewidth]{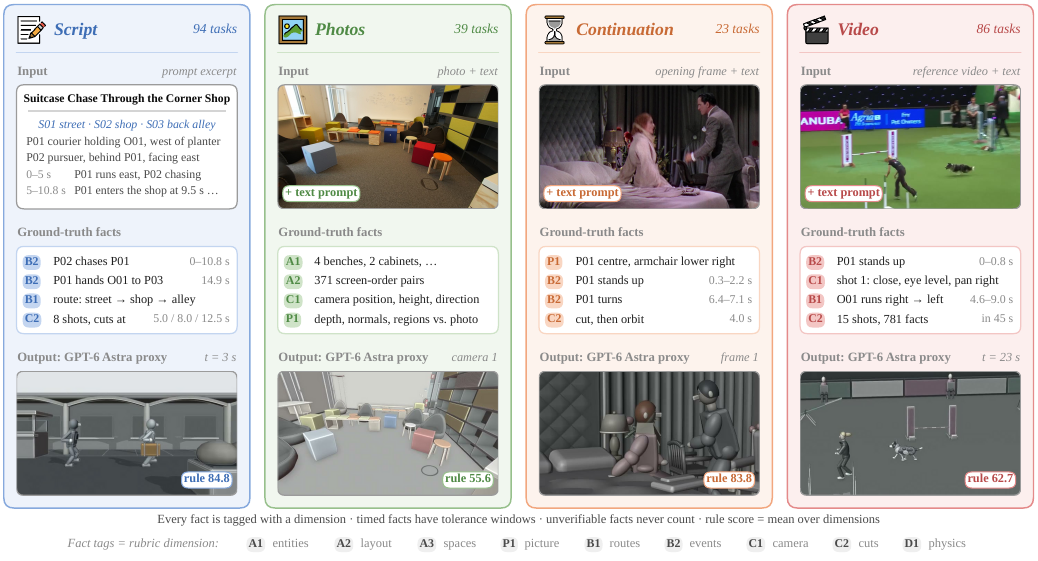}
  \caption{\textbf{The four kinds of input.} One task of each kind. \emph{Top:} what the agent receives (a script excerpt, a photo, an opening frame, a reference video); photo, continuation and video tasks also come with a text brief. \emph{Middle:} some of the hidden GT facts for that task, each tagged with its rubric dimension. \emph{Bottom:} a frame of the proxy video delivered by GPT-6 Astra, with its Rule-based Scene Check score. Timed facts carry tolerance windows, and facts the ground truth cannot verify never count.}
  \label{fig:overview}
\end{figure*}

\Cref{fig:overview} shows one task of each kind, and their prompts are excerpted below.

\paragraph{Script (excerpt).}
{\small\itshape Suitcase Chase Through the Corner Shop. A courier is chased through a shop into a back alley, where a reversing van traps his accomplice\ldots{} S02 Shop: north of the street's east part. A west shelf and an east shelf run north--south either side of the aisle from the front door to the back door\ldots{} P01 person: courier holding O01, west of the planter, facing east. P02 person: pursuer, behind P01, facing east\ldots{} 0--5\,s: P01 runs east with O01, P02 chasing. 5--10.8\,s: P01 enters the shop by the front door at 9.5\,s; P02 follows at 10.8\,s\ldots{} Shots (lens: wide-angle 24\,mm, moderately wide 28\,mm, normal 35\,mm, slightly long 50\,mm)\ldots{} 5--8\,s, hard cut: under the arcade's east end; eye level, normal, facing west; pans right to the door\ldots}

\paragraph{Photos (prompt).}
{\small\itshape Rebuild the place shown in the attached photo as one static 3D proxy blockout, with one camera. Build the place and everything in it at real-world size, and place the camera so that its view matches the photo: position, direction and lens. Output: 12 frames at 24\,fps, 640$\times$426\ldots{} Place: S01 Informal learning lounge. Tag every piece of the place (walls, floor, furniture, fixtures) by class\ldots}

\paragraph{Continuation (excerpt).}
{\small\itshape Continue the attached photo as a proxy previs. Frame 1 must reproduce the photo: the camera's position, direction and lens, and every visible subject and object where it is in the photo\ldots{} Output: 336 frames at 24\,fps (14.00\,s)\ldots{} P01 person: the seated woman beside the bed\ldots{} P02 person: the standing man to her right\ldots{} 0.0\,s: the close eye-level side view of P01 and P02 pulls back in a dolly move. From 0.3 to 2.2 seconds, P01 stands up from the bed\ldots{} 4.0\,s: Cut to a medium eye-level front view of P01 with P02. The camera orbits toward her side\ldots{} From 6.4 to 7.1 seconds, P01 turns back toward P02\ldots}

\paragraph{Reference video (prompt).}
{\small\itshape Rebuild the attached reference video as a proxy previs. Match what the video shows: the places and what stands where, which way things face, how every listed subject moves and what it does when, and the camera of every shot (where it stands, how it frames and moves), with the cuts and transitions at the same times as in the video. Start at the video's first frame. Reference video: 45.02\,s. Output: 1080 frames at 24\,fps (45.00\,s), 640$\times$360. Places: S01 indoor agility arena. Subjects: P01 person: blonde woman with her hair tied in a ponytail\ldots{} O01 dog: medium-sized dog with grey, black, and white fur\ldots}

\subsection{Task statistics}
\label{sec:appstats}
Scripts come in seven writing styles with 9 to 24 scripts each, and photo tasks give a single photo in 14 cases and three to five photos in the other 25. Outputs run about 40~s for scripts and videos, 12~s for continuations and 10~s for photo tasks. A script stages a median of four people in three places over seven shots, and a video four people over 12 shots, with up to 15 people and 38 shots. Scripts and videos carry about 500 and 420 checks at the median, while photo tasks and continuations carry fewer than 100. Beyond events, camera holds 11\% of the checks, routes 8\% and cuts 7\%, layout, picture structure and entities 4 to 5\% each, and spaces and physical validity 2\% each.

\FloatBarrier

\section{Evaluation details}

\subsection{Rule-based Scene Check dimensions}
\label{sec:appdims}
\begin{table*}[!t]
\centering
\caption{\textbf{Rule-based Scene Check dimensions.} The Rule-based Scene Check score is the plain mean of the dimensions a task exercises; P1 applies only to tasks with reference frames (photos, continuations and videos).}
\label{tab:dims}
\footnotesize\renewcommand{\arraystretch}{0.92}
\resizebox{\linewidth}{!}{\begin{tabular}{@{}lp{0.78\linewidth}@{}}
\toprule
Dim. & What is checked against the hidden GT facts\\
\midrule
A1 Entities & every listed subject and prop exists, with the right class and count\\
A2 Layout & relative positions: sides, distances, depth and screen order, against walls\\
A3 Spaces & places, their extents and how they connect\\
P1 Picture structure & per-shot depth, normals, segmentation, edges and facing vs.\ the reference frames\\
B1 Routes & where each subject is over time, the order of places visited, direction\\
B2 Events & whether each event happens, when, in which order, with which posture and gait\\
C1 Camera & per shot: position, height, facing, lens, subjects in frame, camera move\\
C2 Cuts & number of shots, cut times, transitions, shot order and duration\\
D1 Physical validity & no interpenetration, floating or sinking, teleporting, implausible sizes\\
\bottomrule
\end{tabular}}
\end{table*}

\subsection{Rule-based Scene Check scoring}
\label{sec:appscoring}
\paragraph{Gates and replay.} The three gates run 21 automatic checks: the code runs standalone; the video is rendered in full at the specified resolution, frame rate, duration and codec; and the scene reopens in Blender with the task's settings. Fourteen of the 2,662 runs fail a gate and score 0, two of them by Gemini 4 Argon. The replay dumps the tagged entities and places, their oriented boxes and parts, the cameras and the cuts.

\paragraph{Binding and alignment.} Script facts are read in the agent's own frame, whose compass the brief fixes. For media tasks, each place is aligned by a similarity $T(x)=s\,R_\theta x+t$ fitted by weighted least squares to the anchored bodies, with $s\in[0.67,1.5]$. A place with fewer than three anchors is scored by pairwise relations only.

\paragraph{Fact scores.} The 84.4k scorable facts expand into 99.0k checks. A yes-or-no check scores 0 or 1, and most metric checks with error $e$ score $\mathrm{clip}\big((z-e)/(z-f)\big)$, which is full within the registered tolerance $f$ and zero beyond $z$.

\paragraph{Chance rates and physical validity.} The chance rate of a block is $R_b=\frac{1}{200}\sum_{k=1}^{200}P_b(\pi_k S)$, where $\pi_k$ randomly permutes the agent's static objects within each place. Camera checks, which no permutation moves, take $R_b=0$. Physical validity is scaled by $\min(1,E/0.25)$, where $E$ is the mean of layout, routes and events, so that a near-empty scene earns no physical-validity points.

\subsection{VLM judge rubric}
\label{sec:appjudge}
The judge (Gemini~4) watches the proxy video at 2\,fps. Its prompt says that the video comes from a simplified previs, with simple shapes and mannequin-like figures in at most flat colors, so this look is not penalized. It never sees the agent's name or the object names.

\paragraph{Turn 1: blind recap.} Without the brief, the judge describes only what it sees: in time order, who is where and what each person does, with people referred to by their look and objects by their shape (``unidentified shape'' or ``unclear movement'' otherwise); what kind of place it is; and every object it can identify from its shape. For photo tasks it also watches an evaluation-side orbit of the delivered scene (a 360$^\circ$ look from the room center, then a top-down orbit with the ceiling hidden) and lists places and objects only.

\paragraph{Turn 2: verdicts.} The judge then receives the brief (and, for video tasks, the reference video) and checklists generated from the hidden GT facts; items the rules already failed are not asked. It is told that the automatic checks verify presence, positions, layout, camera placement, timing, shot count and physical validity, and that materials, lighting, realism, aesthetics and composition are not judged. Every entry gets \emph{yes}, \emph{partial}, \emph{no} or \emph{n/a} (entity or shot absent, already penalized) with timestamps, and the lower verdict when in doubt.
{\sloppy\begin{description}[style=unboxed,leftmargin=1em,itemsep=1pt,font=\normalfont\bfseries]
  \item[V1 Recognition of shapes and places (key entities, places).] \emph{yes} only if the blind recap already named the category or kind of place; \emph{partial} if it was not named blind but a shape at a plausible spot fits once the brief is known; \emph{no} otherwise.
  \item[V2 Action readability (events).] Would a viewer who has not read the brief describe the footage as that action? \emph{yes}: action and contact are clear; \emph{partial}: the pose changes but the action or contact is unclear; \emph{no}: miming, standing still or another action. Events whose moment is not visible in V3 are left out of V2.
  \item[V3 Visibility at key moments.] At each judged event, and in each shot without events, are the people and objects involved clearly visible? \emph{partial}: small, partly blocked or partly out of frame; \emph{no}: blocked, out of frame, too small, a blank or wall-only frame, the camera inside geometry, flicker or leftover helper objects.
  \item[V4 Story beats.] Does the judge's own blind recap contain each story beat (\emph{yes}: its gist; \emph{partial}: part of it)?
\end{description}}
With \emph{yes} $=1$, \emph{partial} $=\frac12$ and \emph{no} $=0$, each dimension averages its entries (\emph{n/a} left out), and VLM-based Screen Check is 100 times the mean over the applicable dimensions, which on photo tasks are V1 alone.

\subsection{Is the evaluation trustworthy?}
\label{sec:validity}
\paragraph{Determinism and direction.} Rule-based Scene Check is deterministic: each of the 2,786 deliveries that were scored more than once during development received an identical score every time. On 12 high-scoring video deliveries, injected defects lower the Rule-based Scene Check score monotonically and in the intended dimensions: delaying all action by 4~s costs 9.8 points and keeping one person costs 12.1. A 90$^\circ$ camera yaw costs 3.6 points when only the exported cameras are turned, and 21.0 when the ray-cast picture buffers are also re-projected to the turned camera. The physical-validity check misses floating people: lifting them by 1~m costs only 0.8.

\paragraph{Judges.} Re-judged with GPT-6.1 Sol and Qwen3.8 Omni Flash, the agents' VLM-based Screen Check scores correlate with Gemini~4's at Spearman 0.94--0.96, though only at 0.73--0.76 per run; Gemini~4 favors its own family by 1.3 points without changing any rank, and the overall ranking is identical whichever judge is used (Kendall $\tau$ = 1.00). \Cref{tab:judges} lists every pair and the agreement with humans.

\begin{table*}[!htbp]
\centering
\caption{\textbf{Agreement between judges and with humans.} \emph{Agent level}: correlation of the 11 agents' mean VLM-based Screen Check scores; \emph{run level}: Spearman $\rho$ over single runs. Failed runs and the runs GPT-6.1 Sol declined are excluded. Human rows correlate each agent's win rate in the human study with its mean score from that judge. \emph{Mean}: each judge's mean score over all 2,662 runs, failed runs counting 0.}
\label{tab:judges}
\footnotesize\renewcommand{\arraystretch}{0.92}
\setlength{\tabcolsep}{5pt}
\resizebox{\linewidth}{!}{\begin{tabular}{@{}llrccc@{}}
\toprule
 & & & \multicolumn{2}{c}{Agent level} & Run level \\ \cmidrule(lr){4-5}\cmidrule(lr){6-6}
Pair & & Runs & Spearman $\rho$ & Pearson $r$ & Spearman $\rho$ \\
\midrule
Gemini~4 & GPT-6.1 Sol & 2,628 & 0.96 & 0.98 & 0.76 \\
Gemini~4 & Qwen3.8 Omni Flash & 2,648 & 0.94 & 0.97 & 0.73 \\
GPT-6.1 Sol & Qwen3.8 Omni Flash & 2,628 & 0.96 & 0.99 & 0.79 \\
\midrule
Human & Gemini~4 & 242 & 0.92 & 0.94 & -- \\
Human & GPT-6.1 Sol & 242 & 0.92 & 0.91 & -- \\
Human & Qwen3.8 Omni Flash & 242 & 0.90 & 0.93 & -- \\
\midrule
\multicolumn{6}{@{}l}{Mean: Gemini~4 75.3, GPT-6.1 Sol 61.1, Qwen3.8 Omni Flash 61.0} \\
\bottomrule
\end{tabular}}
\end{table*}

\paragraph{Ranking robustness.} Agent means are compared by a paired bootstrap over tasks with 10,000 resamples. The 95\% interval of the score difference includes zero for five of the ten adjacent pairs: GPT-6.1 Sol and Gemini~4 Argon, Gemini~4 Argon and Opus~5.5, GPT-5.6 Sol and DeepSeek V4.1 Flash, Gemini~3.8 Flash and GLM 5.3 Flash, and Doubao Seed 2.1 Pro and Muse Spark 1.3. The 18 scoring variants combine nine scores, namely the overall score with each of the three judges or their mean, the geometric mean of rules and judge, the rules alone and each judge alone, with two weightings, one per task and one per input kind. Their rankings agree with the main one at Kendall $\tau$ of 0.855 or more, and GPT-6 Astra is first under every one of them.

\paragraph{Humans.} On 242 pairs of proxy videos, one per task, an annotator picked the one easier to understand; each agent's win rate correlates with its mean score from each judge at Spearman 0.90--0.92.

\FloatBarrier

\section{Additional results}
\label{sec:appresults}

\subsection{Ablations and capability profiles}
\label{sec:appprofiles}

\begin{table*}[!htbp]
\centering
\caption{\textbf{Ablations} on DeepSeek V4.1 Flash. Each row reruns the agent with one change and reports the paired per-task difference to the main run (mean\,$\pm$\,s.d.). \emph{Rerun}: the same tasks run again unchanged. \emph{Oracle caption}: video tasks with the video replaced by a text description. \emph{No images}: every image returned to the agent is replaced by a text placeholder. $|\Delta|$: mean absolute change of the overall score per task. Cost: agent \$ per task, reference $\rightarrow$ ablation. $^\dagger$The caption is generated from the hidden GT facts (events, cuts, camera), so this row bounds what a perfect perception step would add.}
\label{tab:ablation}
\footnotesize\renewcommand{\arraystretch}{0.92}
\setlength{\tabcolsep}{2.2pt}
\resizebox{\ifdim\width>\linewidth\linewidth\else\width\fi}{!}{\begin{tabular}{@{}llc ccc c c@{}}
\toprule
Change & Tasks & $n$ & $\Delta$Scene & $\Delta$Screen & $\Delta$Overall & $|\Delta|$ & Cost \\
\midrule
Rerun & Script & 8 & $-$2.0{\tiny$\pm$5} & $+$2.5{\tiny$\pm$14} & $+$0.2{\tiny$\pm$7} & 5.5 & 0.57$\rightarrow$0.70 \\
Oracle caption$^\dagger$ & Video & 5 & $+$14.0{\tiny$\pm$8} & $+$17.2{\tiny$\pm$13} & $+$15.6{\tiny$\pm$6} & 15.6 & 1.94$\rightarrow$0.47 \\
No images & Script & 8 & $-$0.1{\tiny$\pm$4} & $-$1.6{\tiny$\pm$16} & $-$0.8{\tiny$\pm$6} & 5.5 & 0.57$\rightarrow$0.59 \\
\bottomrule
\end{tabular}}
\end{table*}

\begin{figure*}[!htbp]
  \centering
  \includegraphics[width=0.95\linewidth]{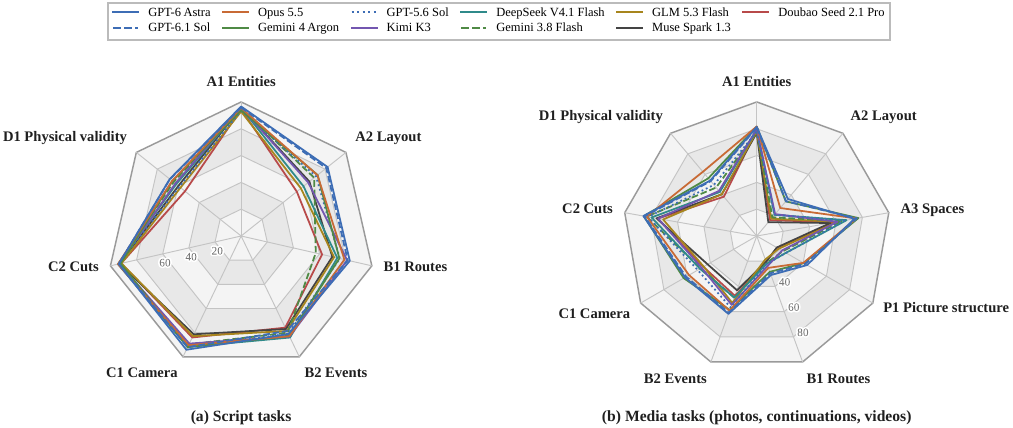}
  \caption{\textbf{Capability profiles of all 11 agents.} Rule-based Scene Check score per dimension (0--100, failed runs count 0) on (a) script tasks and (b) media tasks. Exact values per agent are in \cref{fig:dims}.}
  \label{fig:radar}
\end{figure*}

\begin{figure*}[!htbp]
  \centering
  \includegraphics[width=0.95\linewidth]{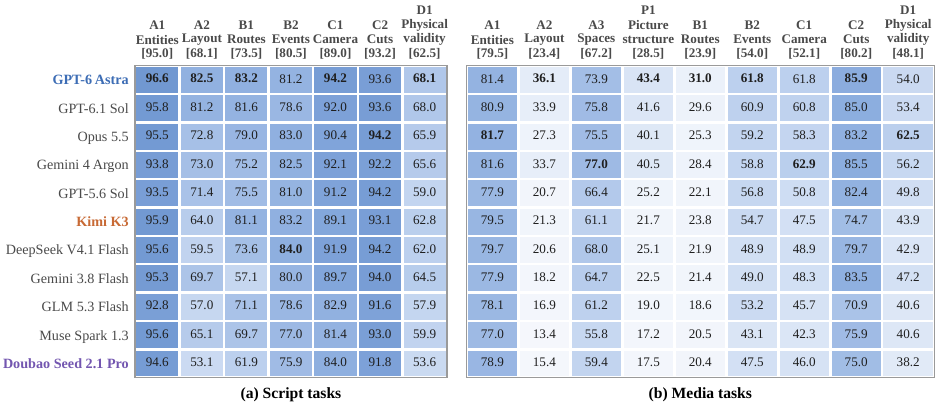}
  \caption{\textbf{Rule-based Scene Check score per dimension} (failed runs count 0; column means in brackets; best per column in bold). (a) Script tasks. (b) Media tasks (photos, continuations, videos). Rows sorted by Rule-based Scene Check score over all tasks.}
  \label{fig:dims}
\end{figure*}

\Cref{fig:radar} profiles all 11 agents across the nine Rule-based Scene Check dimensions, and \cref{fig:dims} gives the exact values, separately for script and media tasks.

\subsection{Where agents improve, and where they do not}
\label{sec:appwhere}
\begin{figure*}[!htbp]
  \centering
  \includegraphics[width=\linewidth]{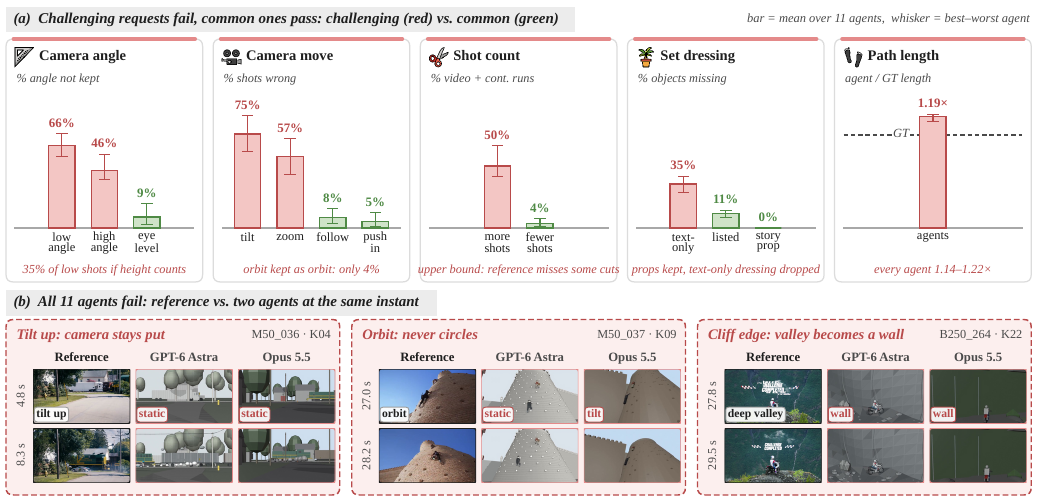}
  \caption{\textbf{Shared failure modes.} (a) Error rates on a challenging request (red) and a common one (green), averaged over the 11 agents; whiskers show the best and worst agent. \emph{Camera angle} and \emph{camera move}: shots whose angle or move type differs from the reference. \emph{Shot count}: video and continuation runs with more or fewer shots than the reference, an upper bound since the reference misses some cuts. \emph{Set dressing}: objects missing from the scene. \emph{Path length}: median ratio of agent path to GT path, pooled over all input kinds. (b) Shots that the rules mark wrong for all 11 agents: the reference frame and two agents at the same instant; tags describe what each one shows.}
  \label{fig:failures}
\end{figure*}

This section details the evidence behind \cref{fig:where,fig:failures} and the observations of \cref{sec:exp}. Unless stated otherwise, gains use the Rule-based Scene Check score, intervals are 95\% paired bootstrap intervals over tasks, and correlations are Spearman $\rho$ over the 11 agents.

\begin{table*}[!htbp]
\centering
\caption{\textbf{Within-family Rule-based Scene Check gains.} Paired over tasks by input kind; 95\% bootstrap intervals, task counts in brackets.}
\label{tab:family}
\footnotesize\renewcommand{\arraystretch}{0.92}
\setlength{\tabcolsep}{3pt}
\resizebox{\linewidth}{!}{\begin{tabular}{@{}lccccc@{}}
\toprule
Step & Script (94) & Photo (39) & Cont. (23) & Video (86) & All (242) \\
\midrule
GPT-5.6 Sol $\rightarrow$ GPT-6.1 Sol & $+$3.6{\tiny\,[2.2, 4.9]} & $+$11.1{\tiny\,[8.8, 13.5]} & $+$5.1{\tiny\,[2.6, 7.6]} & $+$6.9{\tiny\,[5.6, 8.2]} & $+$6.1{\tiny\,[5.2, 7.0]} \\
Gemini 3.8 Flash $\rightarrow$ Gemini 4 Argon & $+$3.4{\tiny\,[0.4, 5.8]} & $+$14.1{\tiny\,[11.4, 17.1]} & $+$7.1{\tiny\,[4.1, 10.5]} & $+$9.7{\tiny\,[8.3, 11.0]} & $+$7.7{\tiny\,[6.3, 9.0]} \\
GPT-6.1 Sol $\rightarrow$ GPT-6 Astra & $+$1.2{\tiny\,[$-$0.2, 2.7]} & $+$0.5{\tiny\,[$-$1.1, 2.1]} & $+$1.7{\tiny\,[$-$0.2, 3.5]} & $+$0.9{\tiny\,[$-$0.3, 2.0]} & $+$1.0{\tiny\,[0.3, 1.8]} \\
\bottomrule
\end{tabular}}
\end{table*}

\paragraph{The spread on visual input is not a difficulty effect.} Within each input kind, we split the tasks at the median of their mean Rule-based Scene Check score. On scripts, the best and worst agents are 12.3 points apart on the harder half and 11.8 on the easier half. The two halves give 18.3 and 22.6 on photos, 18.6 and 13.6 on continuations, and 19.7 and 15.3 on videos. A two-way decomposition splits the Rule-based Scene Check score into agent, task and residual. The agent explains 17\% of the variance on scripts but 21\% on continuations, 28\% on photos and 29\% on videos. The harder script half (mean 76.8) is still easier than the average video task (48.5), so the split controls difficulty within a kind, not across kinds. The spread is a Rule-based Scene Check result. On photos the judge asks only whether subjects and places are recognizable, where agents score 81--94, so photos have the smallest overall-score spread, 13.3.

\paragraph{Generational gains concentrate on visual input.} \Cref{tab:family} gives the gains behind \cref{fig:where}(a). In both families the largest gain on media tasks is in picture structure (P1 $+$16.4 and $+$17.9), followed by layout and camera (A2 $+$13.2 and $+$15.5; C1 $+$10.0 and $+$14.5). The step from GPT-6.1 Sol to GPT-6 Astra is above zero over all tasks but within no single input kind, and every dimension changes by $-$1.9 to $+$2.6. Only two family pairs are available, and the Gemini pair also moves from a Flash to an Argon tier. Under the Qwen3.8 Omni Flash judge, Gemini 4 Argon also gains 11.2 VLM-based Screen Check points on scripts, so the concentration on visual input holds for Rule-based Scene Check.

\paragraph{Picture structure forms two tiers.} Per task, the mean P1 of the top four agents exceeds that of the other seven on 39 of 39 photo tasks, by 38.3 points on average (95\% CI 34.6--41.8). The same holds on 11 of 11 continuation tasks (25.0 points) and on 70 of 71 video tasks, where every agent stays at 12--26 and the gap is only 9.6. Photo tasks give no camera intrinsics, so agents must estimate position, direction and lens. Entities and layout barely separate the agents on photos (A1 7.6, A2 8.1). No agent exceeds 26 on layout: better agents match the view, while none generates an accurate layout of the room behind it. The top four are also the newest and most expensive agents, so the tiers do not isolate which capability crosses the threshold. Split by dimension as weighted in the score, the 15.0-point Rule-based Scene Check gap between the top four and the other seven on photos comes 45\% from picture structure, 18\% from camera, 14\% from physical validity, 13\% from spaces, 6\% from layout and 3\% from entities.

\begin{figure*}[!htbp]
  \centering
  \includegraphics[width=\linewidth]{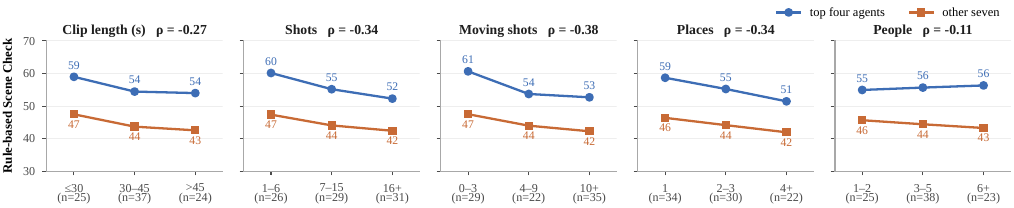}
  \caption{\textbf{What makes a reference video hard.} The 86 video tasks fall into three bins along five properties of the reference clip, task counts under each bin. Lines give the mean Rule-based Scene Check score of the top four agents by overall score and of the other seven; $\rho$ is the Spearman correlation between the property and the task's mean score over all 11 agents. Failed runs count 0.}
  \label{fig:difficulty}
\end{figure*}

\paragraph{Reading the reference limits video tasks.} \Cref{fig:difficulty} splits the 86 video tasks by properties of the reference clip. Scores fall with clip length, shots, moving shots and places (Spearman $-$0.27 to $-$0.38), but hardly with the number of people or objects ($-$0.11 each). On shots, moving shots and places, the top four lose 7--8 points from the first to the last bin and the other seven about 5, so complex clips narrow the gap between them rather than widen it. The oracle caption of \cref{tab:ablation} raises the score on all five tasks at a quarter of the cost, gaining in routes, events, camera and cuts but not in layout. The caption comes from the hidden GT facts and covers one agent, so it bounds what perfect perception would add.

\paragraph{Photos rank the agents differently.} The overall-score rankings of scripts, continuations and videos agree at Kendall $\tau$ = 0.85--0.96, while photos agree with each of them at only 0.56--0.67. Opus 5.5 leads Rule-based Scene Check on photos, 4.0 points above GPT-6 Astra (95\% CI 1.6--6.5) and higher on 28 of 39 tasks. Its lead comes mainly from physical validity (D1 53.6 vs.\ 41.4) and camera (C1 71.7 vs.\ 66.3). With only 39 tasks, the photo ranking rests on a small sample.

\paragraph{Error rates.} \Cref{fig:where}(c) reads errors from the per-check results of Rule-based Scene Check, over delivered runs. A check counts as an error when it scores below 0.5. For each run, a rate is the share of its applicable checks of one type that are errors, and each agent's rate is the mean over its runs. Picture regions and picture depth are the region-layout and depth checks of picture structure. Framing, camera angle, camera move and camera height check each shot's shot size, pitch class, move type and camera height against the hidden GT facts. Missing objects counts the objects in the hidden GT facts that the scene lacks. More shots than ref.\ is the share of video and continuation runs with more shots than the reference.

\paragraph{Shared errors barely change with capability.} In \cref{fig:where}(c), errors in reading the picture fall steadily from the weakest to the best agent, while the shared errors stay within a few points. With 11 agents a weak $\rho$ does not prove that no relation exists, so the claim rests on the narrow ranges and on the best and weakest agents being close. Over-cutting depends on the measure. The share of runs with more shots than the reference gives $\rho$ = $-$0.18. The share of video runs with at least one extra cut gives $-$0.48 (52--86\%, highest for GLM 5.3 Flash). Weaker agents thus add more cuts, while the best still adds them in 55\% of video runs. Path excess holds on scripts and videos. The median ratio to the ground truth is 1.20 on scripts (7,307 paths; 1.15--1.23 per agent), 0.70 on continuations (138 paths) and 1.16 on videos (1,379 paths; 0.96--1.42 per agent). Scripts hold 83\% of the paths, so they set the pooled 1.19 of \cref{fig:failures}(a). Script paths in the hidden GT facts average 49 samples, and only 4\% have two or fewer, so the excess is not an artifact of straight-line references. Direction counts cases whose ratio to the reference exceeds 1.5 or falls below two-thirds: shot counts over 1,191 video and continuation runs, path lengths per path and scene extents per task.

\paragraph{Challenging and common requests.} \Cref{fig:failures}(a) pairs each challenging request with a common counterpart scored the same way. Camera angles use the scorer's reading from camera pitch, as in Rule-based Scene Check; counting camera height as well, 35\% of low-angle shots come out at eye level. Orbits are classified by the move type of the agent's camera, which keeps 4\% as orbits and turns 47\% into sideways moves. The tilt shot of \cref{fig:failures}(b) fails for all 11 agents, and nine of them keep the camera still. The pairs were chosen after the errors were seen, and that agents fall back on common choices because challenging requests are scarce in their training data is our reading of the error direction, not a tested cause.

\paragraph{Looking does not fix.} Each agent's runs are split into thirds by the images it views, ties broken by task id, and a run's score is taken relative to the other agents' mean on the same task, which removes task difficulty. Picture structure is treated the same way on the media tasks. Without this control the third that looks most scores 1.5 points lower at the median, since hard tasks draw more looking and score lower. Image counts include every image returned to the agent after its first Blender call, so on photo, continuation and video tasks they include reference frames viewed again. In the no-image ablation of \cref{tab:ablation}, all eight runs received zero images; it covers one mid-ranked agent and eight script tasks. These comparisons describe associations rather than the effect of looking, since agents may well look more when a task goes wrong.

\paragraph{Verdicts against errors.} We take the text an agent writes after each batch of images it views once it has made its first Blender call. A run counts as claiming a match when this text states that the render matches the reference, brief or photo, or that it looks right or correct, by a fixed pattern list. Per-shot error rates come from the Rule-based Scene Check items behind \cref{fig:where}(c). A regression with agent and task fixed effects and a bootstrap over tasks compares runs with and without a claim. Camera-angle errors differ by $-$0.8 points (95\% CI $-$4.0 to $+$2.1), camera moves by $+$0.1 and missing events by $-$0.6. Framing differs by $-$2.1 ($-$3.7 to $-$0.5) and shot count by $-$2.5 ($-$5.7 to $+$0.7). A stricter pattern list and Claude Code runs alone leave camera angle, camera move and events within 1.5 points, while Claude Code runs alone show 4.7 fewer shot-count errors. The Codex CLI keeps only reasoning summaries, so GPT agents claim a match in 123 of 717 runs, and most claims come from the other eight agents. The analysis is correlational, and the pattern list is not hand-checked.

\paragraph{Failed Blender calls.} For each Blender call that ends in an error, we take the last line of its traceback. Of 3,796 failed calls, 36\% use an attribute, keyword or enum value that the Blender~5.2 API does not have; errors from the GPU-less sandbox's renderer are not counted as such. \texttt{Action.fcurves}, which Blender~5 replaced with layered actions, accounts for 23\%. A run counts as calling it if any of its calls raises it. With agent and task fixed effects, runs with an API failure score $+$0.1 points (95\% CI $-$0.8 to $+$1.0) against runs without, and runs calling \texttt{Action.fcurves} $-$0.3 ($-$1.1 to $+$0.9), so later rounds repair these calls.

\paragraph{Build rhythm.} Agents spread their Blender calls differently over a run, measured in tool calls and pooled over each agent's runs. Gemini 4 Argon and Opus 5.5 leave 61\% and 54\% of their Blender calls to the last third, whereas GPT-6 Astra and GLM 5.3 Flash leave only 35\% and 37\%, and both rhythms reach the top. In Claude Code, Opus 5.5 runs Blender 6 times per task and DeepSeek V4.1 Flash 24, yet Opus 5.5 scores 4.5 points higher.

\subsection{Event completion and legibility}
\label{sec:appmore}

\begin{figure}[!htbp]
  \centering
  \includegraphics[width=0.84\linewidth]{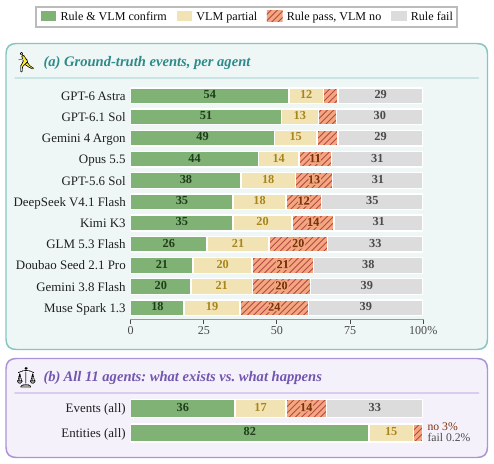}
  \caption{\textbf{Event completion.} Each hidden GT fact is triangulated by both scorers: the rules pass it and the VLM confirms it on screen, confirms it partly, or rejects it; or the rules already fail it. Facts without a usable VLM verdict are excluded. (a) Events per agent: confirmed events fall from 54\% (GPT-6 Astra) to 18\% (Muse Spark 1.3), and weaker agents more often stage an event the rules accept but the screen does not show. (b) Pooled over agents, entities almost never fail, while events do.}
  \label{fig:events}
\end{figure}

\paragraph{Correct is not always legible, and events separate agents most.}
The VLM judge is asked only about facts the rules accept. Of those, about a third of the actions cannot be read from the render (68\% legible on average, from 86\% for GPT-6 Astra to 50\% for Muse Spark 1.3), and a blind recap retells only 55\% of the story. Weaker agents thus err more and also make their correct parts harder to read. Chaining the two scores per event (\cref{fig:events}), only 36\% of annotated events are both accepted by the rules and confirmed on screen: about half for the four strongest agents (44--54\%) and a fifth for the four weakest, the widest gap of any error type. A further 31\% pass the rules but are confirmed only partly (17\%) or rejected (14\%) by the judge, so the rules alone would overstate event completion. Conversely, the rules miss some posture events such as falls, which lowers the confirmed share.

\subsection{Results by scene domain and screen dimension}
\label{sec:appdomains}
\paragraph{Scene domains.} \Cref{tab:domains} breaks the overall score down by the seven scene domains. GPT-6 Astra leads six of them and ties GPT-6.1 Sol on Homes at 78.3. Each domain's ranking agrees with the overall ranking at Kendall $\tau$ = 0.85--0.96, with the same top four in all seven, so the ranking does not hinge on the domain. Streets are the weakest domain for all 11 agents.

\paragraph{Screen dimensions.} \Cref{tab:dimsall} splits VLM-based Screen Check into its four dimensions. Story is the lowest dimension for all 11 agents, while recognition varies only from 80.9 to 91.7; \cref{sec:appmore} analyzes how legible the actions are.

\begin{table*}[!t]
\centering
\caption{\textbf{Overall score by scene domain.} Mean of $(\mathrm{Scene}+\mathrm{Screen})/2$ over the tasks of each of the seven domains of \cref{fig:stats}, task counts in parentheses. Per column, the best score is in \textbf{bold}. $^{\blacktriangle}$ and $^{\blacktriangledown}$ mark domains where an agent scores at least 2 points above or below its own \emph{All}.}
\label{tab:domains}
\footnotesize\renewcommand{\arraystretch}{0.92}
\setlength{\tabcolsep}{4pt}
\resizebox{\linewidth}{!}{\begin{tabular}{@{}lrrrrrrrr@{}}
\toprule
Agent & Homes & Offices & Streets & Shops & Industry & Sports & Nature & All \\
 & \scriptsize(34) & \scriptsize(27) & \scriptsize(42) & \scriptsize(19) & \scriptsize(31) & \scriptsize(46) & \scriptsize(43) & \scriptsize(242) \\
\midrule
GPT-6 Astra & \textbf{78.3}\phantom{$^{\blacktriangle}$} & \textbf{78.9}\phantom{$^{\blacktriangle}$} & \textbf{74.4}$^{\blacktriangledown}$ & \textbf{80.7}$^{\blacktriangle}$ & \textbf{77.0}\phantom{$^{\blacktriangle}$} & \textbf{79.2}\phantom{$^{\blacktriangle}$} & \textbf{77.3}\phantom{$^{\blacktriangle}$} & \textbf{77.7} \\
GPT-6.1 Sol & \textbf{78.3}$^{\blacktriangle}$ & 76.5\phantom{$^{\blacktriangle}$} & 72.4$^{\blacktriangledown}$ & 77.3\phantom{$^{\blacktriangle}$} & 76.7\phantom{$^{\blacktriangle}$} & 76.5\phantom{$^{\blacktriangle}$} & 76.0\phantom{$^{\blacktriangle}$} & 76.1 \\
Gemini 4 Argon & 75.4\phantom{$^{\blacktriangle}$} & 74.9\phantom{$^{\blacktriangle}$} & 73.6\phantom{$^{\blacktriangle}$} & 76.4\phantom{$^{\blacktriangle}$} & 75.1\phantom{$^{\blacktriangle}$} & 75.3\phantom{$^{\blacktriangle}$} & 75.1\phantom{$^{\blacktriangle}$} & 75.0 \\
Opus 5.5 & 76.6$^{\blacktriangle}$ & 75.9\phantom{$^{\blacktriangle}$} & 70.3$^{\blacktriangledown}$ & 74.4\phantom{$^{\blacktriangle}$} & 75.2\phantom{$^{\blacktriangle}$} & 74.4\phantom{$^{\blacktriangle}$} & 73.4\phantom{$^{\blacktriangle}$} & 74.1 \\
GPT-5.6 Sol & 70.4\phantom{$^{\blacktriangle}$} & 70.5\phantom{$^{\blacktriangle}$} & 66.9$^{\blacktriangledown}$ & 72.8$^{\blacktriangle}$ & 70.9\phantom{$^{\blacktriangle}$} & 70.5\phantom{$^{\blacktriangle}$} & 71.1\phantom{$^{\blacktriangle}$} & 70.2 \\
DeepSeek V4.1 Flash & 68.4\phantom{$^{\blacktriangle}$} & 69.5\phantom{$^{\blacktriangle}$} & 65.9$^{\blacktriangledown}$ & 71.6\phantom{$^{\blacktriangle}$} & 71.6$^{\blacktriangle}$ & 70.8\phantom{$^{\blacktriangle}$} & 70.5\phantom{$^{\blacktriangle}$} & 69.6 \\
Kimi K3 & 66.4\phantom{$^{\blacktriangle}$} & 66.7\phantom{$^{\blacktriangle}$} & 63.6$^{\blacktriangledown}$ & 67.1\phantom{$^{\blacktriangle}$} & 69.5\phantom{$^{\blacktriangle}$} & 69.4\phantom{$^{\blacktriangle}$} & 69.3\phantom{$^{\blacktriangle}$} & 67.5 \\
Gemini 3.8 Flash & 66.1\phantom{$^{\blacktriangle}$} & 64.5\phantom{$^{\blacktriangle}$} & 59.8$^{\blacktriangledown}$ & 66.3\phantom{$^{\blacktriangle}$} & 64.1\phantom{$^{\blacktriangle}$} & 67.1$^{\blacktriangle}$ & 65.7\phantom{$^{\blacktriangle}$} & 64.7 \\
GLM 5.3 Flash & 60.5$^{\blacktriangledown}$ & 64.6\phantom{$^{\blacktriangle}$} & 59.5$^{\blacktriangledown}$ & 65.4\phantom{$^{\blacktriangle}$} & 61.5$^{\blacktriangledown}$ & 67.1$^{\blacktriangle}$ & 67.1$^{\blacktriangle}$ & 63.7 \\
Doubao Seed 2.1 Pro & 62.4\phantom{$^{\blacktriangle}$} & 63.7$^{\blacktriangle}$ & 57.6$^{\blacktriangledown}$ & 63.4$^{\blacktriangle}$ & 62.1\phantom{$^{\blacktriangle}$} & 62.8\phantom{$^{\blacktriangle}$} & 59.3\phantom{$^{\blacktriangle}$} & 61.3 \\
Muse Spark 1.3 & 61.2\phantom{$^{\blacktriangle}$} & 62.9\phantom{$^{\blacktriangle}$} & 56.3$^{\blacktriangledown}$ & 64.3$^{\blacktriangle}$ & 58.2$^{\blacktriangledown}$ & 63.8$^{\blacktriangle}$ & 62.3\phantom{$^{\blacktriangle}$} & 61.1 \\
\bottomrule
\end{tabular}}
\end{table*}

\begin{table*}[!t]
\centering
\caption{\textbf{VLM-based Screen Check per dimension} (Gemini~4 judge). Each dimension is the mean over the tasks that exercise it, task counts in parentheses; failed runs count 0. \emph{Screen} is the score of \cref{tab:main}. Per column, the best score is in \textbf{bold}. \Cref{fig:dims} gives the Rule-based Scene Check dimensions.}
\label{tab:dimsall}
\footnotesize\renewcommand{\arraystretch}{0.92}
\setlength{\tabcolsep}{5pt}
\resizebox{\linewidth}{!}{\begin{tabular}{@{}lrrrr|r@{}}
\toprule
Agent & V1 Recognition & V2 Action & V3 Visibility & V4 Story & Screen \\
 & \scriptsize(242) & \scriptsize(202) & \scriptsize(203) & \scriptsize(202) & \scriptsize(242) \\
\midrule
GPT-6 Astra & \textbf{91.7} & \textbf{85.6} & \textbf{88.7} & \textbf{68.6} & \textbf{85.4} \\
GPT-6.1 Sol & 91.1 & 82.3 & 86.3 & 64.4 & 83.1 \\
Gemini 4 Argon & 87.9 & 81.2 & 83.7 & 66.8 & 81.7 \\
Opus 5.5 & 88.8 & 76.7 & 82.9 & 63.1 & 79.8 \\
GPT-5.6 Sol & 89.9 & 69.7 & 80.3 & 58.1 & 77.5 \\
DeepSeek V4.1 Flash & 88.8 & 72.1 & 82.4 & 57.2 & 77.9 \\
Kimi K3 & 85.5 & 67.5 & 77.5 & 55.7 & 73.5 \\
Gemini 3.8 Flash & 83.5 & 55.6 & 74.2 & 46.0 & 68.7 \\
GLM 5.3 Flash & 84.3 & 57.9 & 76.0 & 46.6 & 69.5 \\
Doubao Seed 2.1 Pro & 81.3 & 54.0 & 71.0 & 41.4 & 66.2 \\
Muse Spark 1.3 & 80.9 & 50.2 & 68.5 & 41.2 & 65.0 \\
\bottomrule
\end{tabular}}
\end{table*}

\subsection{Do proxy errors reach the real-world video?}

\label{sec:downstream}
\Cref{fig:proxy} gives the real-world video generation model Seedance~2.5 the proxy videos of a strong and a weak agent on two video tasks, with one prompt shared by both clips of a panel. In both panels, the real-world video generated from the weak agent's proxy keeps its error: the high camera and flat lot in (a), and the distant car in (b). Camera errors can thus reach the real-world video. These pairs show that errors can reach real-world videos, not how often they do. Correlating scores with real-world video quality across all agents is left to future work.

\subsection{Case gallery}
\label{sec:appgallery}
\begin{figure*}[!p]
  \centering
  \includegraphics[width=\linewidth]{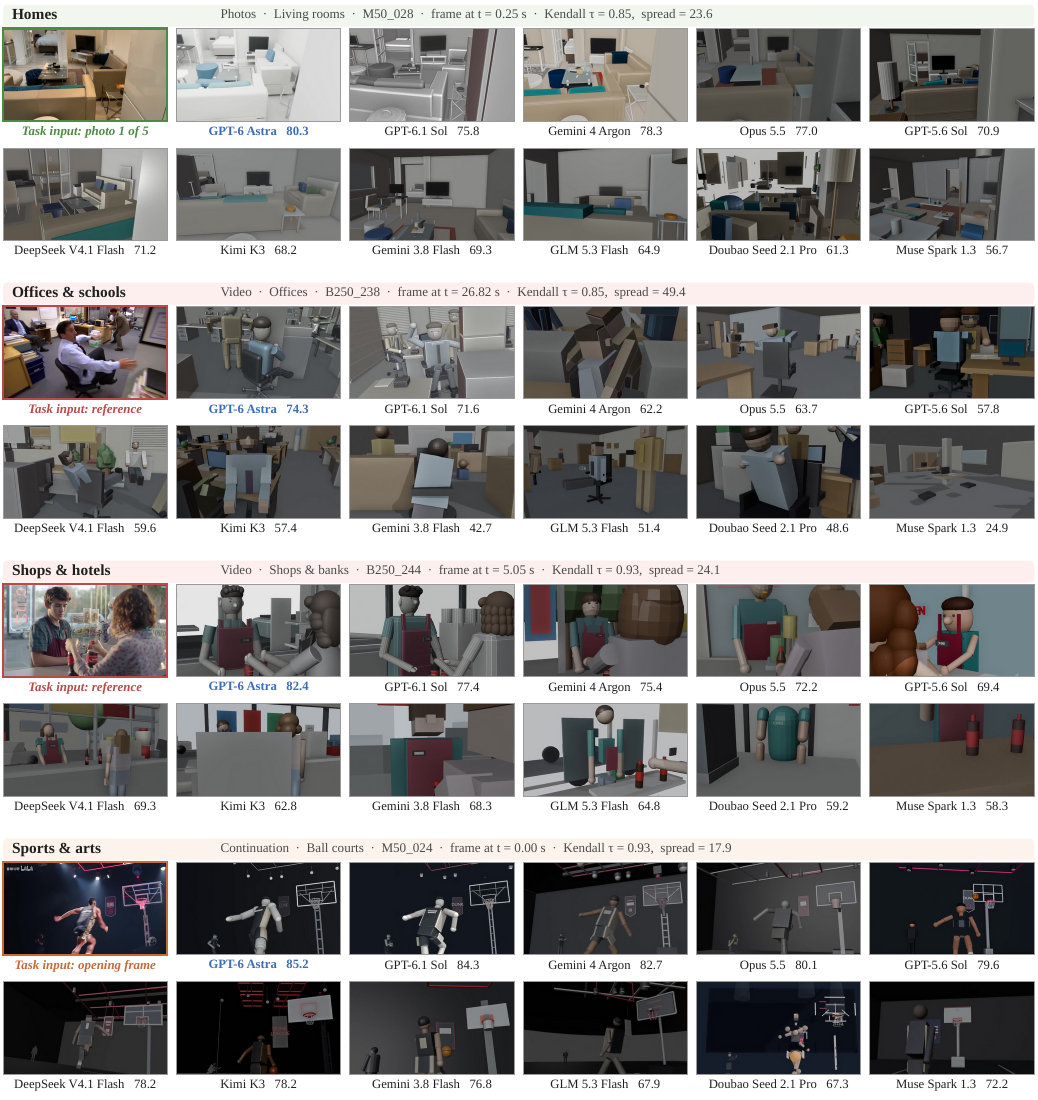}
  \caption{\textbf{Case gallery: homes, offices and schools, shops and hotels, sports and arts.} Each domain shows the task input, a reference-video frame, the first of five photos or the opening frame of a continuation, then the frame at the same moment from each agent's proxy video, agents in the order of \cref{tab:main}. Each frame carries the agent's overall score on that task, the best per task in bold. Each task has the highest Kendall $\tau$ between its agent ranking and the overall ranking among the media tasks of its domain on which all 11 runs succeed, ties broken by the larger score spread. Frame times are reference keyframes with people or action in view.}
  \label{fig:gallery}
\end{figure*}

\begin{figure*}[!p]
  \centering
  \includegraphics[width=\linewidth]{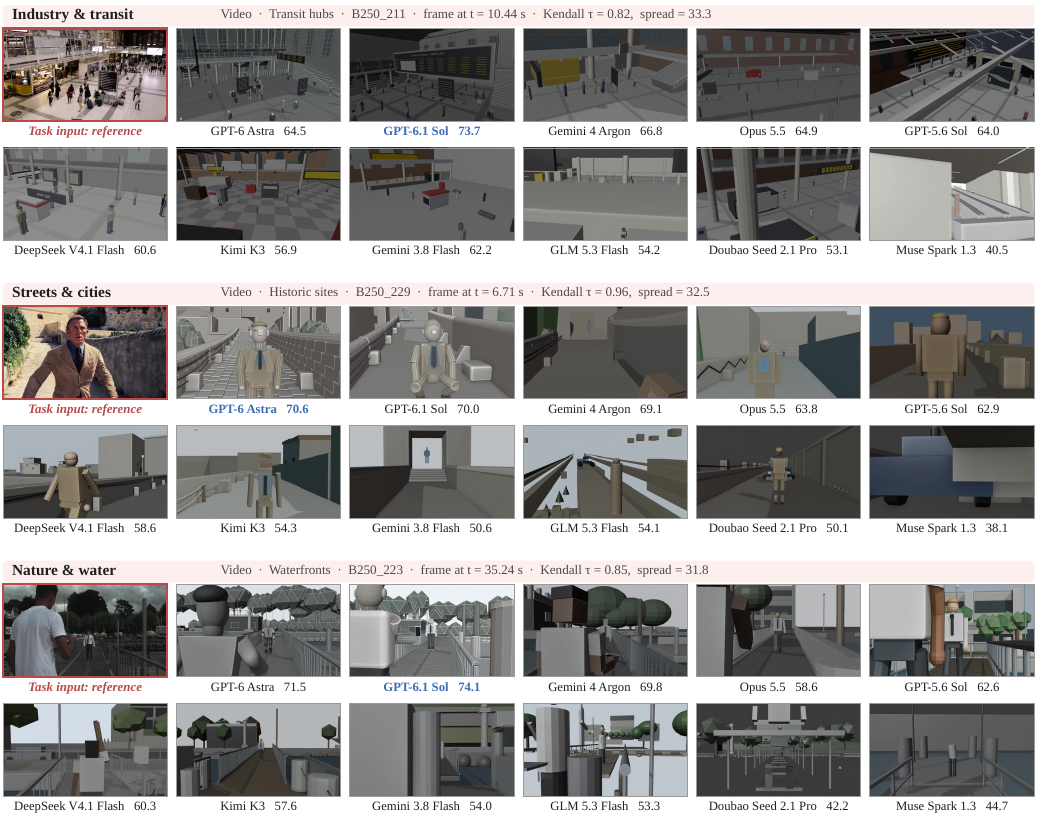}
  \caption{\textbf{Case gallery: industry and transit, streets and cities, nature and water.} Layout and task selection as in \cref{fig:gallery}.}
  \label{fig:gallery2}
\end{figure*}

\Cref{fig:gallery,fig:gallery2} show one task per scene domain, with the proxy video of every agent at the moment of the reference frame.

\begin{figure}[!htbp]
  \centering
  \includegraphics[width=0.92\linewidth]{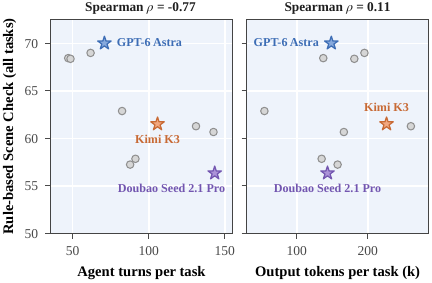}
  \caption{\textbf{Effort vs.\ score.} Mean turns and output tokens per task against the Rule-based Scene Check score (all tasks), one marker per agent (stars: best, median and weakest); Spearman $\rho$ over the 11 agents.}
  \label{fig:effort}
\end{figure}

\subsection{Agent statistics and cost}
\Cref{tab:process} gives per-task means of the agents' process and cost over all 242 tasks. \Cref{fig:effort} relates effort to score: across the 11 agents, more turns go with a lower Rule-based Scene Check score (Spearman $\rho$ = $-$0.77), while output tokens do not track it ($\rho$ = 0.11).

\begin{table*}[!htbp]
\centering
\caption{\textbf{Process and cost per task on all 242 tasks.} Means of turns, tool calls, wall-clock minutes, input tokens (millions, cache hits included) and output tokens (thousands, reasoning included); cost in US dollars at the official list prices of late October 2026, computed per request from the API relay logs. \emph{Failed}: runs without a valid scene after re-execution, counted as in \cref{tab:main}. Agents in the order of \cref{tab:main}.}
\label{tab:process}
\footnotesize\renewcommand{\arraystretch}{0.92}
\setlength{\tabcolsep}{5pt}
\resizebox{\linewidth}{!}{\begin{tabular}{@{}lrrrrrrr@{}}
\toprule
Agent & Turns & Tools & Min. & In (M) & Out (k) & Cost (\$) & Failed \\
\midrule
GPT-6 Astra & 71 & 70 & 93 & 11.2 & 149 & 32.59 & 0 \\
GPT-6.1 Sol & 62 & 60 & 105 & 10.7 & 195 & 6.71 & 0 \\
Gemini 4 Argon & 49 & 85 & 73 & 5.8 & 181 & 4.45 & 2 \\
Opus 5.5 & 47 & 61 & 54 & 6.7 & 137 & 7.82 & 0 \\
GPT-5.6 Sol & 82 & 81 & 39 & 6.8 & 54 & 6.19 & 0 \\
DeepSeek V4.1 Flash & 131 & 141 & 91 & 27.9 & 261 & 1.89 & 0 \\
Kimi K3 & 106 & 129 & 196 & 20.6 & 226 & 17.66 & 4 \\
Gemini 3.8 Flash & 143 & 141 & 88 & 16.0 & 166 & 2.73 & 0 \\
GLM 5.3 Flash & 91 & 116 & 89 & 12.7 & 135 & 0.73 & 3 \\
Doubao Seed 2.1 Pro & 143 & 168 & 116 & 22.8 & 143 & 5.96 & 1 \\
Muse Spark 1.3 & 88 & 130 & 108 & 7.8 & 157 & 3.84 & 4 \\
\bottomrule
\end{tabular}}
\end{table*}

\FloatBarrier

\end{document}